\pdfoutput=1
\documentclass[11pt]{article}
\usepackage{acl}
\usepackage{mathtools}
\usepackage{graphicx}
\usepackage{wasysym}
\usepackage{times}
\usepackage{latexsym}
\usepackage[T1]{fontenc}
\usepackage[utf8]{inputenc}
\usepackage{microtype}
\usepackage{inconsolata}
\usepackage{tabularx}
\usepackage{arydshln}
\usepackage{amsfonts}
\usepackage{bm}
\usepackage{bbm}
\usepackage{subcaption}
\usepackage{booktabs}
\usepackage{multirow}
\usepackage{makecell}
\usepackage{xcolor}
\usepackage{colortbl}
\usepackage{tikz}
\usepackage{xspace}
\usepackage{array,multirow}
\usepackage{pgfplots}
\usepackage{verbatim}
\usepackage{fancyvrb}
\usepackage{caption}
\usepackage{tcolorbox}
\pgfplotsset{compat=1.18}
\usepackage{amssymb}
\usepackage[ruled,linesnumbered,commentsnumbered]{algorithm2e}
\usepackage{graphicx}

\definecolor{mygreen}{RGB}{46,139,87}
\definecolor{iyellow}{RGB}{255,250,205}
\definecolor{ipurple}{RGB}{230,230,250}
\definecolor{myred}{RGB}{238,44,44}
\definecolor{myblue}{RGB}{30,144,255}
\definecolor{myorange}{RGB}{255,127,80}
\definecolor{mypurple}{RGB}{255,20,147}
\definecolor{lcolor}{rgb}{0.56, 0.0, 1.0}

\def\hot{\raisebox{-0.55ex}{\includegraphics[width=0.7em]{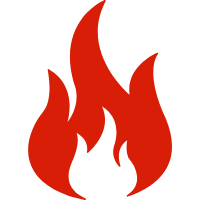}}}
\def\froze{\raisebox{-0.55ex}{\includegraphics[width=0.7em]{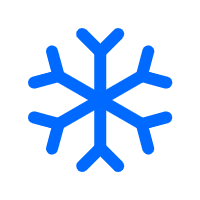}}}

\title{Self-Powered LLM Modality Expansion for Large Speech-Text Models}

\author{
Tengfei Yu$^{1}$~~
 Xuebo Liu$^{1}$\thanks{~~Corresponding Author}~~
 Zhiyi Hou$^{2}$~~
 Liang Ding$^{3}$~~
    \bf{Dacheng Tao}$^{4}$~~
    \bf{Min Zhang}$^{1}$\\
    \textsuperscript{\rm1}Institute of Computing and Intelligence, Harbin Institute of Technology, Shenzhen, China \\
    \textsuperscript{\rm2}Faculty of Computing, Harbin Institute of Technology, Harbin, China \\
    \textsuperscript{\rm3}The University of Sydney~~~
    \textsuperscript{\rm4}Nanyang Technological University\\
    \texttt{\{tengfeiyu,zhiyihou\}@stu.hit.edu.cn, \{liuxuebo,zhangmin2021\}@hit.edu.cn} \\
    \texttt{liangding.liam@gmail.com, dacheng.tao@ntu.edu.sg} \\}
    
\begin{document}
\maketitle
\begin{abstract}
Large language models (LLMs) exhibit remarkable performance across diverse tasks, indicating their potential for expansion into large speech-text models (LSMs) by integrating speech capabilities. Although unified speech-text pre-training and multimodal data instruction-tuning offer considerable benefits, these methods generally entail significant resource demands and tend to overfit specific tasks.
This study aims to refine the use of speech datasets for LSM training by addressing the limitations of vanilla instruction tuning. We explore the instruction-following dynamics within LSMs, identifying a critical issue termed \textit{speech anchor bias}—a tendency for LSMs to over-rely on speech inputs, mistakenly interpreting the entire speech modality as directives, thereby neglecting textual instructions.
To counteract this bias, we introduce a self-powered LSM that leverages augmented automatic speech recognition data generated by the model itself for more effective instruction tuning. Our experiments across a range of speech-based tasks demonstrate that self-powered LSM mitigates speech anchor bias and improves the fusion of speech and text modalities in LSMs. Data, code and scripts are freely available at \url{https://github.com/ytf-philp/Self-powered-LSM}.
\end{abstract}

\section{Introduction}
Recent advances in large language models (LLMs) \citep{gpt3,llama,palm2} have achieved remarkable performance across numerous tasks~\citep{gpt4}. To further extend the capabilities of LLMs, many efforts~\cite{liu2023visual,zhang2023video,rubenstein2023audiopalm} have been made to integrate LLMs with multi-modal encoders, aiming to endow models with multimodal capabilities. Speech, as a typical modality, shares similar semantics with text and has become a widely studied area.

To expand the speech capabilities of LLMs, several studies adopt multitask learning \cite{chu2023qwen} or unified speech-text pre-training \cite{wang2023viola,rubenstein2023audiopalm}. While yielding notable performance, these methods require extensive computational resources. To address this, \citet{tang2023salmonn} construct multimodal speech instruction datasets to fine-tune LLMs for developing large speech-text models (LSMs). However, vanilla datasets often reveal LSMs' deficiencies in following instructions \cite{wang2023blsp,pan2023cosmic}. To avoid task-specific overfitting and enhance generalizability, careful development of datasets and methodologies is necessary.

Given the prevalence of automatic speech recognition (ASR) datasets, using this data to build multimodal speech instruction datasets is straightforward. Thus, we question whether LSMs can be fully developed solely by collecting this type of training data. To achieve this, this study first explores the attention dynamics between speech inputs and instructions, identifying inherent training challenges presented by vanilla ASR training.
Our analysis reveals a pronounced tendency for LSMs trained with speech instructional data to overly concentrate on speech, neglecting instructions. This predisposition causes LSMs to mistakenly interpret the entire speech modality as instruction, limiting the model's responses to scenarios that closely mirror its training data. We define this phenomenon as \textit{speech anchor bias}.

Motivated by this finding, we argue that emphasizing the role of instructions during the training process is crucial for enhancing the instruction-following ability of LSMs. To achieve this, we introduce a novel self-powered LSM that utilizes augmented instructional data generated by the model itself to facilitate the expansion of modality capabilities. 
Experiments conducted on ASR, speech translation (ST), speech-language understanding (SLU), and question answering (QA) demonstrate that our method, even when applied solely to ASR datasets, effectively improves LSM performance across diverse tasks, thereby alleviating speech anchor bias and enhancing the integration of speech and text modalities in LSMs

The \textbf{main contributions} of this paper are:
\begin{itemize}
\item We unveil a critical issue in LSM training—\textit{speech anchor bias}, where direct instruction tuning causes excessive reliance on speech inputs, diminishing the model's competence with instructions.
\item We introduce an innovative self-powered LSM, which leverages self-generated data to rapidly enhance the speech modality capabilities of LLMs.
\item Further analysis confirms that our method effectively reduces speech anchor bias, achieving better alignment between speech and text.
\item We publicly release our self-powered augmentation dataset, hoping it will benefit further research within the community.
\end{itemize}
\section{Related Work}
\subsection{Pretraining LSMs for Modality Extension}
Recent advancements in pretraining LSMs can be divided into two categories. The first focuses on self-supervised speech representation learning without text supervision, enabling unsupervised learning from extensive unlabeled speech data \citep{mohamed2022self,wu2023speechgen}. This approach also facilitates the generation of high-quality speech tokens \citep{lakhotia2021generative,popuri2022enhanced,zhang2023speechtokenizer}. The second approach unifies speech and text within a single model, as exemplified by models like VioLA \cite{wang2023viola}, AudioPalm \cite{rubenstein2023audiopalm}, SpeechGPT \cite{zhang2023speechgpt}, LauraGPT \cite{chen2023lauragpt}, and Qwen-Audio \cite{chu2023qwen}. These models leverage a shared vocabulary for speech and text, or use a multi-task training framework to continue training the LLM with a speech dataset, thereby enhancing their capabilities across various speech tasks. However, pretraining an LSM requires more resources and complex procedures, making it difficult to quickly adapt to different LLM backbones and rendering it a challenging endeavor.

\subsection{Expand LLMs with Speech Capabilities}
Following the advancement of ChatGPT, AudioGPT \cite{huang2023udiogpt} has enabled LLMs to process speech by interacting within task-specific model pipelines. Efforts towards end-to-end integration include aligning speech and text embeddings through connection modules between speech encoders and LLMs \cite{chen2023xllm,wu2023decoder,yuwenyi,wang2023blsp} or integrating LoRA \cite{hu2021lora} into LLMs, thereby enhancing their speech capabilities. Notably, studies by \citet{tang2023salmonn} and \citet{wang2023blsp} indicate that directly training LLMs with both speech and target transcripts can lead to overfitting on specific speech tasks, potentially causing modality imbalance in LSM. Despite efforts to address this issue, previous work still requires task-specific training adjustments, limiting efficient adaptation across various LLMs. Our work comprehends the model's behavior through layer-wise attention and introduces a training framework to efficiently expand LLMs with speech capabilities.

\subsection{Self-distillation for LLM Finetuning}
\begin{figure}[tbp]
    \centering
    \includegraphics[width=0.36\textwidth, height=0.16\textheight]{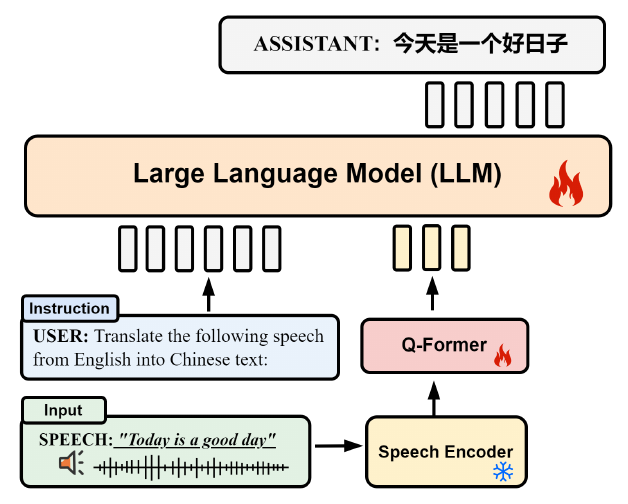}
    \caption{Model architecture of LSM. }
    \label{fig:all}
\end{figure}
The utilization of self-distillation datasets in training LLMs has recently emerged as a significant area of interest. Self-Instruct \cite{wang2023self} and WizardLM \cite{xu2024wizardlm} leverage generated responses from LLMs to conduct supervised fine-tuning, while approaches like Self-Refine \cite{madaan2024self} and Self-Reward \cite{yuan2024self} use these responses as iterative feedback to enhance output quality. Additionally, \citet{yang2024self} employ self-distilled responses from LLMs to address distribution gaps and combat catastrophic forgetting during fine-tuning. Diverging from these studies, our paper ventures into the multi-modality field, employing instructions as the primary driving mechanism and significantly augmenting the speech modality capabilities of LLMs.

\section{Methodology}
\subsection{Preliminary}

\paragraph{Model Architecture}
The model architecture of LSM is shown in Figure~\ref{fig:all}. 
We use the encoder component of Whisper-small \citep{radford2023robust} as the speech encoder and employ Vicuna-7B-1.5 \citep{chiang2023vicuna} as the large language model. 
Q-Former \citep{li2023blip}, serving as the connection module, employs $N$ trainable queries $\mathbf{Q}$ within its stacked blocks. The output sequence, integrated with the text instruction, is then fed into the LLM to generate the text response.

\paragraph{Training Objective}
We consider a set of model parameters parameterized by $\theta$ for training. Let $\mathbf{s}$ represent the speech input and $\mathbf{t}$ its corresponding target text sentence. The whisper encoder receives the original sequence $\mathbf{s}$ as input and produces its contextual representation $\mathbf{H}$. To equip the Q-Former with the ability to handle variable-length general speech inputs, we use the window-level segment strategy \citep{tang2023salmonn} to segment $\mathbf{H}$ into $L$-sized window representations and utilize Q-Former at the window level to output textual tokens $\mathbf{Z}$. The log-likelihood training objective for the parallel speech-text pair  $\mathbf{(s, t)}$ is:

\begin{footnotesize}
\begin{equation}
    \label{eq:final-loss1}
    \begin{split}
        \hat{\theta} &= \arg\min_{\theta} \left(-\log P(\mathbf{t} | \mathbf{s}, \mathbf{i}; \theta)\right)
                \\
        &= \arg\min_{\theta} \left(-\sum_{m=1}^M \log P(t_m | \mathbf{t}_{<m}, \mathbf{s}, \mathbf{i}; \theta)\right),
    \end{split}
\end{equation}
\end{footnotesize}
where $M$ is the length of the target text, $t_m$ is the $m$-th target token, and $\mathbf{i}$ is the text embedding of instruction. The loss is a standard causal language modeling loss, which predicts the next token based on prior tokens. We use the same prompt template as vicuna \citep{chiang2023vicuna} to keep training consistent and do not compute the loss for the instruction during training \citep{wang2023blsp}.

\subsection{Modality Imbalance for LSMs}
\label{understand}
\begin{figure}[t]
    \centering
    \includegraphics[width=0.48\textwidth, height=0.16\textheight]{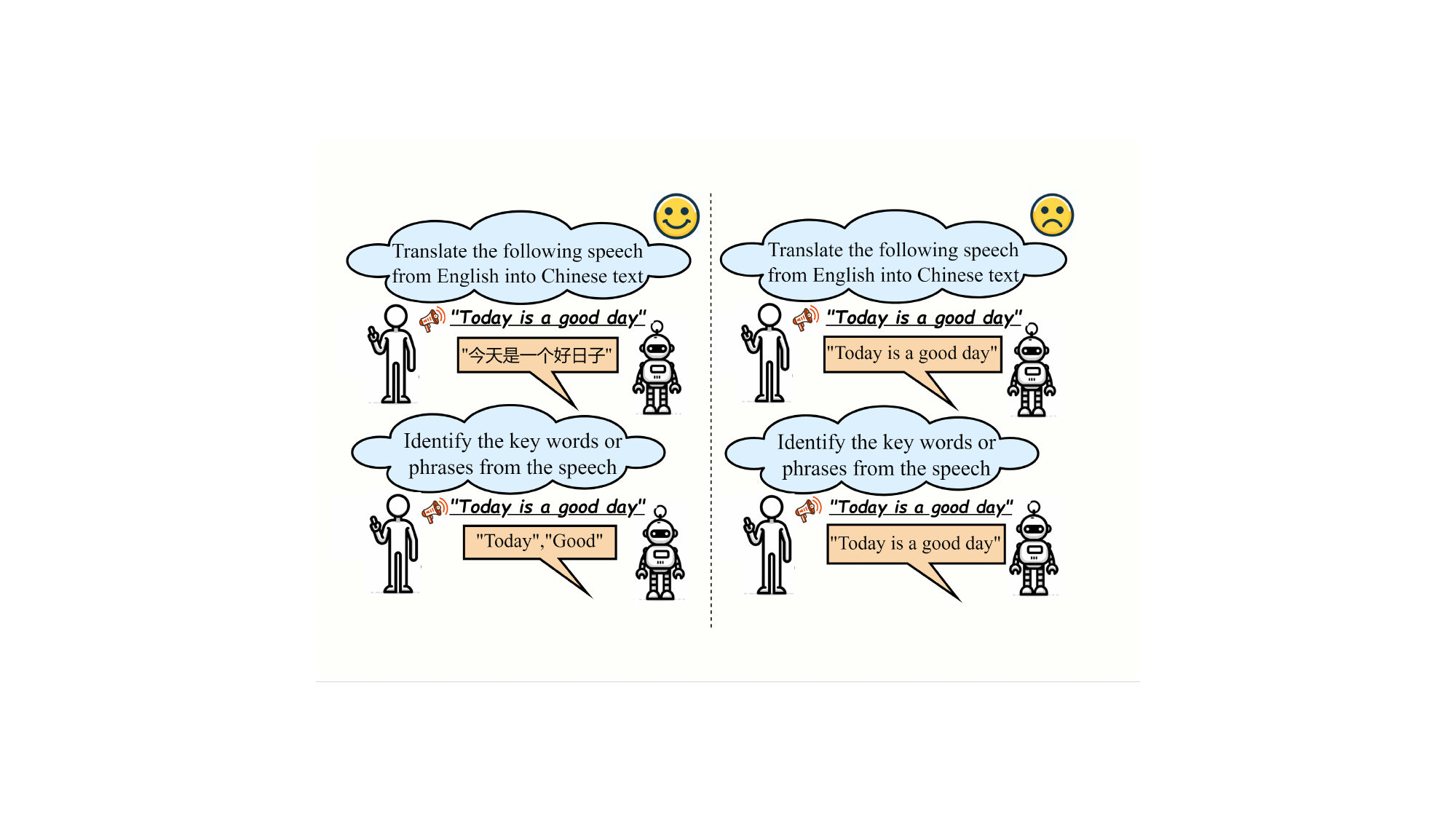}
    \caption{ The left shows a well-trained LSM should possess the capability to follow instructions, whereas the right displays directly fine-tuned model with speech instructional data does not enable the acquisition of speech modality expansion capability.}
    \label{fig:intro}
\end{figure}

To enhance the speech capabilities of LLMs, a common approach involves fine-tuning these models using multimodal instructional samples and organizing data into a structured format that includes textual instructions, speech inputs, and text responses. However, existing research \cite{tang2023salmonn,wang2023blsp} has highlighted a significant deficiency in LSM's ability to adhere to instructions when trained with vanilla speech instructional data. We illustrate this challenge in Figure~\ref{fig:intro}. To investigate the reason behind this modality imbalance during LSM training, we conduct analysis by examining the attention mechanisms across both speech and instructional inputs, exploring information interaction.

\paragraph{Attention Comparision}
\begin{figure*}[t]
    \centering
    \includegraphics[width=0.83\textwidth, height=0.18\textheight]{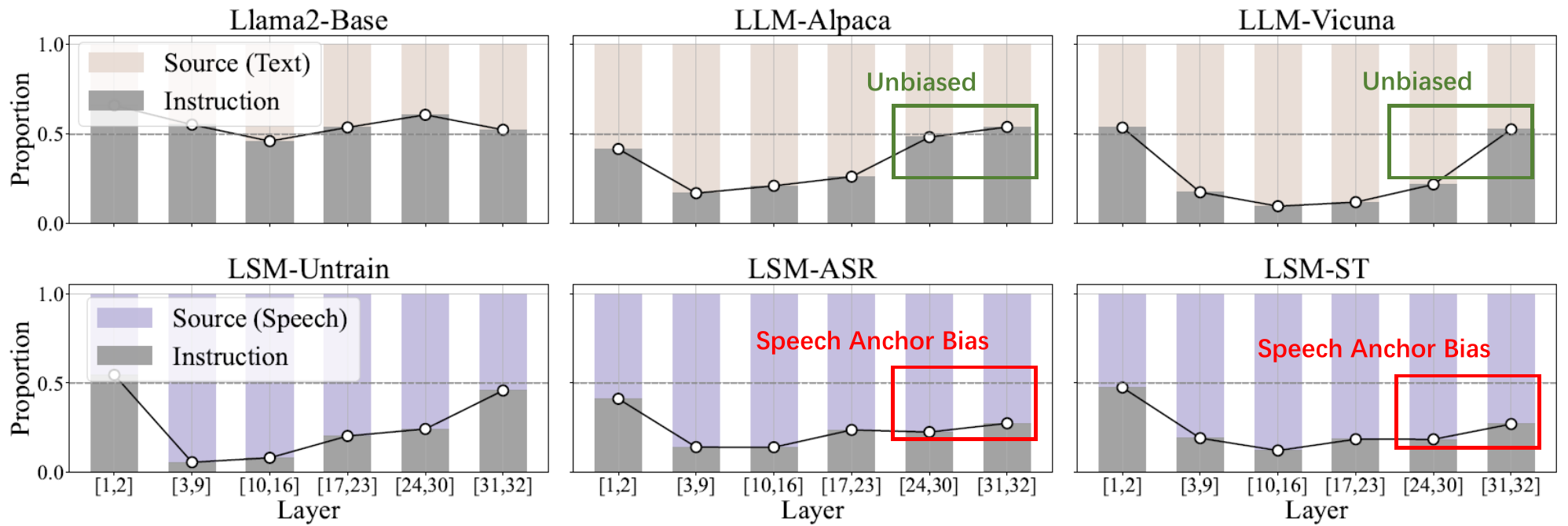}
    \caption{The comparison of the layer-wise behavior in instruction-following LLM versus instruction-ignoring LSM.``Source'' refers to text input for LLMs, whereas denotes speech input for LSM. As the layer deepens, the proportion of instructions diminishes in LSM while increasing in LLM. The red borders show that LSMs excessively focus on speech representations and ignore instructions. }
    \label{fig3:attention}
\end{figure*}
\begin{figure*}[htbp]
    \centering
    \includegraphics[scale=0.48]{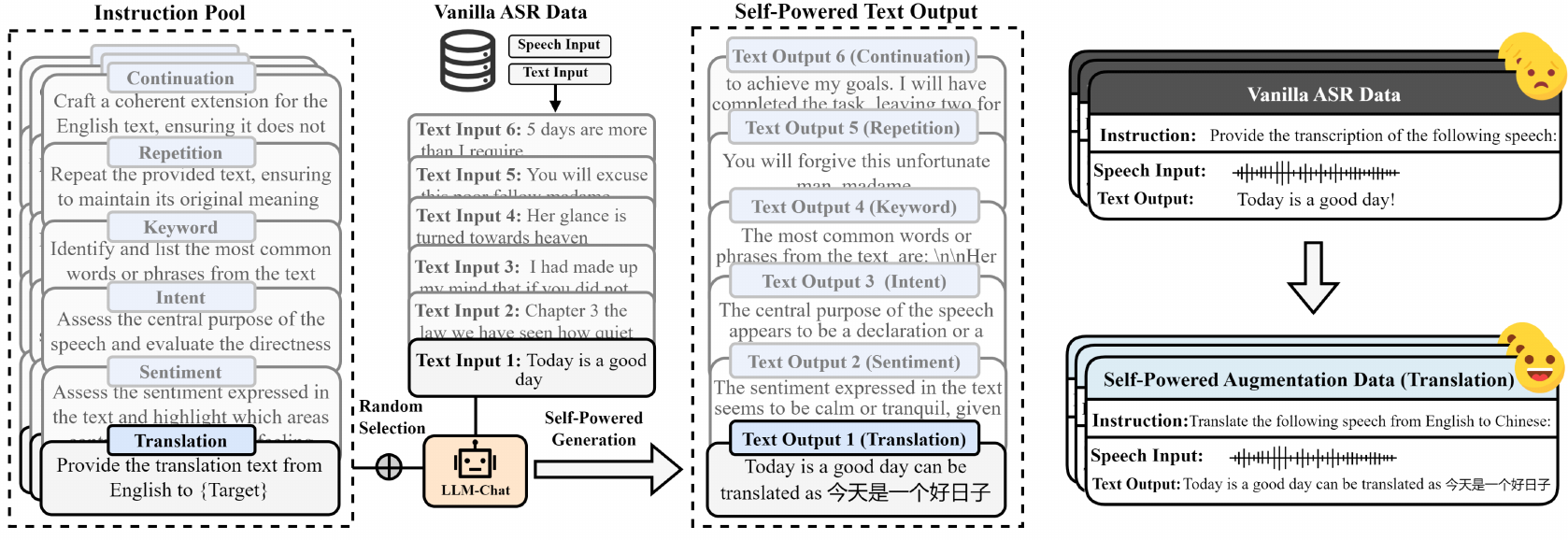}
    \caption{Process of self-powered data augmentation: Self-Powered data is generated by prompting the LLM with instructions alongside the text from the vanilla ASR dataset. The self-powered data is then used to train the LSM. }
    \label{fig4:main}
\end{figure*}

Attention weights is a common interpretation tool that aims to discover the inherent patterns in the attention interaction to ascertain which input vectors contribute~\citep{clark19,liu2020understanding}. 
For multi-modal inputs, the features themselves also affect critical token interactions. 
Meanwhile, numerous studies argue that analyzing the attention mechanism with only attention weights overlooks the effect of the transformed vector \citep{wiegreffe2019attention,bibal2022attention}. 
To gain a more in-depth analysis of attention interaction, we refer the \textit{norm of the weighted transformed vector} \citep{kobayashi2020attention} and define the metric $\mathbf{A}_j$\footnote{The more details of the definition are provided in Appendix~\ref{Formulation}.}denoting the average attention score from the $j$-th input token to the model's output\footnote{During the generation of the $m$-th token, decoder-only model appends the preceding $m-1$ tokens to the end of the input sequence as part of its next token prediction process. To simplify our analysis, we focus on computing the scores exclusively for the tokens within the initial input's length when generating the $m$-th token.}.

Inspired by~\citet{wang2023label}, we introduce quantitative metrics derived from $\mathbf{A}$ to elucidate the information flow from speech inputs and text instructions to the generated sequence. These metrics are expressed as:\par
\begin{footnotesize}
\begin{equation}
\begin{split}
    \mathrm{S_{Instruction}} = \frac{1}{\mathbf{|I|}}\sum_{\mathbf{I}_j\in \mathbf{I}}{\mathbf{A}_j},  ~~
    \mathrm{S_{speech}} = \frac{1}{\mathbf{|Z|}}\sum_{\mathbf{Z}_j\in \mathbf{Z}}{\mathbf{A}_j} ,
\end{split}
\end{equation}
\end{footnotesize}
where $\mathrm{ S_{instruction}}$ and $\mathrm{S_{speech}}$ mean the information flow from instruction part $\mathbf{I}$ and speech part $\mathbf{Z}$ to output sequence.
To assess the relative focus on instruction versus speech components within input features, we calculate the proportion $\eta$ for each layer as $\frac{\mathrm{S_{Instruction}}}{\mathrm{S_{Instruction}}+\mathrm{S_{Speech}}}$.
The $\eta$ value indicates the relative contributions of speech and instruction to model predictions, offering insights into the model's attention allocation during generation.
\paragraph{Differences between LLMs and LSMs}
By using quantitative metrics, we conducted a comparative analysis of layer-wise behavior between different types of instruction-tuning in LLMs and LSMs regarding how they handle instruction and source inputs. Detailed training and evaluation information is available in Appendix~\ref{setup}. Figure~\ref{fig3:attention} shows that after training, LLMs such as Alpaca and Vicuna exhibit dynamic attention shifts: the early layers distribute attention evenly between instructions and source inputs; the middle layers prioritize source information; and the deeper layers refocus on instructions, integrating them into the outputs. In contrast, LSMs, after vanilla training, especially those trained on ASR and ST tasks like LSM-ASR and LSM-ST, \textit{consistently favor speech inputs over instructions across all layers}.

\paragraph{Speech Anchor Bias} Although LSMs are inherently designed to process dual-modal inputs—speech and text—our understanding reveals a significant bias toward speech inputs during training, resulting in the model's complete oversight of textual instructions. During inference, this tendency causes LSMs to interpret the entire speech modality as the instructions that appeared in the training stage, limiting the model's responses to more diverse text instructions. We term this issue ``speech anchor bias'', which is the main reason for causing modality imbalance when using LLMs to expand speech capacity for LSMs.

\subsection{Modality Expansion via Self-Powered Augmentation}

Building on insights from \S~\ref{understand}, it is imperative to avert modality imbalance during LSM training by emphasizing the role of instructions in the training process. To achieve this, we propose self-powered LSM, which capitalizes on self-powered data to enhance the speech modality capabilities of LLMs for LSMs. A comprehensive process of our proposed method is illustrated in Figure~\ref{fig4:main}.

\paragraph{Self-Powered Data Generation} 

Given an LSM $\theta$, we delineate it into three components: the speech encoder $\theta_{s}$, the Q-former $\theta_q$, and LLM $\theta_{l}$. We organize an instruction pool $\mathcal{I}$ into $K$ distinct tasks ${\mathcal{I}_1, \mathcal{I}_2, \ldots, \mathcal{I}_K}$. Each task is associated with $m$ specific types of instances. For each speech-text paired ASR data $\mathbf{(s,t)}$, we select a task from a pool of tasks with equal probability, then randomly employ a specific instruction $\mathbf{i}_{div}$ from $m$ instances. This instruction is used along with the text $\mathbf{t}$ to prompt $\theta_{l}$, resulting in the generation of a self-powered text output $\mathbf{\hat{t}}$, as described by the following equation:\par
\begin{footnotesize}
\begin{equation}
{\mathbf{\hat{t}}} = \arg \max_{\hat{\mathbf{t}}}P(\mathbf{\hat{t}}|\mathbf{t}, {\mathbf{i}_{div}}; \theta_{l}) .
\end{equation}
\end{footnotesize}
Subsequently, the vanilla ASR training sample $(\mathbf{s}, \mathbf{t})$ is supplanted with the newly generated self-powered augmentation sample $(\mathbf{s}, \mathbf{\hat{t}}, \mathbf{i}_{div})$. 
It is important to highlight that any form of instruction instance can contribute to the generation of self-powered text.
To diversify and specify the types of instructions, we craft distinctive categories within the speech domain and generate multiple instances of instructions for each category.

\paragraph{Self-Powered Model Training} 
We use constructed self-powered augmentation data for LSM training. During training, the speech encoder is frozen to ensure the stability of speech features while the Q-former and LLM are fine-tuned for alignment. The training objective can be formulated as: \par
\vspace{-3.5mm}
\begin{footnotesize}
\begin{equation}
\hat{\theta}_q,\hat{\theta}_l = \arg\min_{(\theta_q,\theta_l)}\left(-\log P(\hat{\mathbf{t}}|\mathbf{s}, \mathbf{i}_{div}; \theta^{\froze}_s,\theta^{\hot}_q,\theta^{\hot}_l)\right) .
\end{equation}
\end{footnotesize}
\vspace{-3.5mm}\par
By employing self-powered augmentation data, we can significantly enhance the model's focus on instructions during the training process. In the subsequent section, we provide a theoretical exposition of how self-powered augmentation data effectively avoids modality imbalance during LSM training.

\subsection{Discussion}
\label{sec:4.2}
When training LSMs with vanilla speech instructional data, speech anchor bias causes models to disregard textual instructions, potentially skewing the distribution $P(\mathbf{t}|\mathbf{s}, \mathbf{i}; \theta)$ to closely align with $P(\mathbf{t}|\mathbf{s}; \theta)$. Then the training objective is degraded into $ -\log P(\mathbf{t}|\mathbf{s};\theta)$, restricting the model's response to the tasks within the training dataset.

Conversely, training LSMs with self-powered augmentation data effectively redirects the training objectives to maximize the likelihood of an nstruction-constrained distribution and then strategically shifts the model’s focus toward closer alignment with the instruction $\mathbf{i}$.\par
\vspace{-3.5mm}
\begin{footnotesize}
\begin{align}
     \mathcal{L}(\theta) 
&= -\log P(\mathbf{t}|\mathbf{s}, \mathbf{i}; \theta) \quad \notag \\
    & \doteq -\log P(\mathbf{t}|\mathbf{s}; \theta) \quad\quad\quad \text{\textit{(Speech Anchor Bias)}}\notag \\
    & \doteq -\log P(\mathbf{\hat{t}}|\mathbf{s}; \theta) \quad\quad\text{\textit{(Self-Powered Generation)}} \notag\\
    &\doteq -\log P(M(\mathbf{t, i})|\mathbf{s}; \theta) \quad  \text{\textit{   
 (Modified Objective)}} \notag\\
    &\text{where } M(\mathbf{t, i}) =  \arg \max_{\mathbf{\hat{t}}}  P(\mathbf{\hat{t}|t, i; \theta}). \notag
\end{align}
\end{footnotesize}
\vspace{-3.5mm}\par
\noindent As a result, self-powered model training can improve the model's responsiveness to instructional inputs and effectively mitigate speech anchor bias.

\section{Experiment}
\begin{table}[t]
\centering
\scalebox{0.77}{
\begin{tabular}{lccc}
\toprule
\bf Task & \bf Label Type & \bf \#Num. &\bf \#Samples \\
\midrule
Speech Recognition & Ground-truth & 5 & 237,480 \\  
Content Repetition  & Self-Powered &5 & 237,471  \\
Intent Recognition & Self-Powered & 5 &237,428  \\
Sentiment Analysis  & Self-Powered & 5 & 237,434 \\
Keyword Extract & Self-Powered & 5 & 237,353 \\
Continuation & Self-Powered & 5 &  237,391\\
Speech Translation & Self-Powered & (5,5,5,5) & 237,466 \\
\bottomrule
\end{tabular}}
\caption{Statistic of training dataset. ``\#Num.'' refers to the number of instructions per task. We use the En-\{De, Zh, Fr, Es\} tasks for speech translation. }
\label{tab:statistic}
\end{table}
\subsection{Training Data}
We trained the LSM using several well-established ASR datasets: LibriSpeech-960h \cite{panayotov2015librispeech}, GigaSpeech-L \cite{chen2021gigaspeech}, and Common Voice 4.0 \cite{ardila2019common}, totaling 4,500 hours of training data. Six task types were employed to construct the instruction pool. GPT-4 \cite{gpt4} was utilized to generate unique instructional instances for each task, which were then used to generate self-powered data from the ASR datasets. Additionally, we randomly selected 230k samples from the vanilla ASR dataset and integrated them with the self-powered data for model training. Table~\ref{tab:statistic} provides detailed statistics of the training data.
\begin{table*}[htbp]
\centering
\renewcommand{\arraystretch}{1.1}
\scalebox{0.72}{
\begin{tabular}{llcccccccccc}
\toprule
\multirow{2}{*}{\bf ID~~Method} &\multirow{2}{*}{\bf \#Para} & \multicolumn{2}{c}{\bf ASR$\downarrow$} &  \multicolumn{2}{c}{\bf ST$\uparrow$} &\bf ER$\uparrow$ &\multicolumn{2}{c}{\bf KE$\uparrow$} & \bf IC$\uparrow$ & \multicolumn{2}{c}{\bf QA$\uparrow$}   \\
\cmidrule(lr){3-4} \cmidrule(lr){5-6}  \cmidrule(lr){7-7}  \cmidrule(lr){8-9}  \cmidrule(lr){10-10} \cmidrule(lr){11-12}
& & Clean & Other  & CoVoST & MuSTC & MELD & Light & Water & FSC & WebQ & BoolQ  \\
\midrule
\multicolumn{12}{c}{\it Existing Method} \\
1~~Qwen-Audio\textsuperscript{*} \cite{chu2023qwen} & 8.4B    &\bf 2.6 & \bf 5.1 & \bf 24.5  & 22.0 & 49.1 &0.4&1.1&38.4 & 71.4 & 12.2 \\
2~~BLSP\textsuperscript{*} \citep{wang2023blsp} & 6.9B  & 16.8  &22.3 & 7.5 & 14.7 & 32.1 & 3.0 & \bf 57.7 & 60.8 &  70.0 & 60.4  \\
\multicolumn{12}{c}{\it Implemented Method} \\
3~~Vanilla IT  & 6.9B    & 7.8 & 13.1 & 0.2 & 0.0  &0.0 & 0.0 & 0.0 & 0.0 & 44.3 & 0.0 \\
4~~BLSP & 6.9B  & 26.6  &35.6 & 8.8 & 14.6 & 30.0 & 4.6 & 10.3 & 41.9 &  71.1 & 33.6   \\
\multicolumn{12}{c}{\it Our Method with BackBone Model: Vicuna-7B-1.5 } \\

5~~Self-Powered LSM \footnotesize{with Whisper-small} &6.9B  & 3.8 &  8.0  &17.5 & 21.4 & 51.4 &30.7 & 34.9 & 47.2 & 72.4 & 59.8  \\

6~~Self-Powered LSM \footnotesize{with Whisper-medium} & 7.1B  & 4.0 & 7.9 & 19.2  & 23.7 &  49.6  & 32.3 & 42.5 &58.2 &  73.4 & 60.4   \\
7~~Self-Powered LSM \footnotesize{with Whisper-large} & 7.4B  &  3.3 &  6.1 &   20.7 & \bf 23.8 & \bf 51.5 & \bf 36.2& 47.1 & \bf 61.0 & \bf 73.6 & \bf 61.4    \\
\bottomrule
\end{tabular}}
\caption{Results span all tasks. ``Small'', ``Medium'', and ``Large'' refer to the sizes of the whisper encoders equipped on our method. ``\textsuperscript{*}'' means we test the open-sources 7B model with the same instruction. The ST outcomes represent an average from En-to-X. More details of ST results are provided in Appendix~\ref{appendix:main}.}
\label{tab:main_all}
\end{table*}

\subsection{Training Setup}
Self-powered LSM employs the encoder part of Whisper \cite{radford2023robust} model as the speech
encoder, and Vicuna-7B-1.5 \cite{chiang2023vicuna} as the backbone LLM.
In the Q-former block, we set $N = 1$ for a single trainable query and $L = 17$ to represent approximately $0.33$ seconds per window. 
We freeze the parameters of the speech encoder and train the Q-former and LLM using a batch size of $512$, a learning rate of $2e-5$, a weight decay of $0.05$, and warmup steps totaling $100$. All the models are trained on eight 80GB A800 GPUs across $2$ epochs.

\begin{table*}[htbp]
\centering
\renewcommand{\arraystretch}{1.1}
\scalebox{0.77}{
\begin{tabular}{lcccccccccc}
\toprule
\multirow{2}{*}{\bf Method} & \multicolumn{2}{c}{\bf ASR$\downarrow$} &  \multicolumn{2}{c}{\bf ST$\uparrow$} &\bf ER$\uparrow$ &\multicolumn{2}{c}{\bf KE$\uparrow$} & \bf IC$\uparrow$ & \multicolumn{2}{c}{\bf QA$\uparrow$}   \\
\cmidrule(lr){2-3} \cmidrule(lr){4-5}  \cmidrule(lr){6-6}  \cmidrule(lr){7-8}  \cmidrule(lr){9-9} \cmidrule(lr){10-11}
& Clean & Other  & CoVoST & MuST-C & MELD & Light & Water & FSC & WebQ & BoolQ  \\
\midrule
Self-Powered LSM  & \bf 3.8 & \bf 8.0  & \bf 17.5 & \bf 21.4 & \bf 51.4 & \bf 30.7 & \bf 34.9 & 47.2 & \bf 72.4 & \bf 59.8  \\
~~~~~~~ w/o LLM FT  & 8.5 & 14.5 & 16.8 & 20.2 & 35.4 &  6.9 & 5.9 &  \bf 76.2 & 68.5 & 44.4 \\
~~~~~~~ w/ LoRA FT   &  16.1 & 24.3  &  11.3 &  16.0  &  36.8 &  25.4 &  56.7 & 39.0 & 70.8 & 58.8  \\
\bottomrule
\end{tabular}}
\caption{Results span all tasks with different training strategies.}
\label{tab:strategy}
\end{table*}
\subsection{Evaluation}
For ASR, we evaluated the model using the LibriSpeech test-clean and test-other sets, adopting word error rate (WER) as the performance metric. The efficacy of ST was assessed on the CoVoST \cite{wang2020covost} test set for English (En)-\{Chinese (Zh), German (De)\} translations, and the MuST-C \cite{di2019must} tst-COMMON dataset for En-\{French (Fr), Spanish (Es)\} translations, employing the SacreBLEU score \citep{post2018call} as the evaluation.
For SLU, we evaluated the model’s capabilities through several tasks. Emotion recognition (ER) was assessed on the MELD dataset \cite{poria2018meld} using the Micro-F1 score. Keyword extraction (KE) was evaluated using the SNIP Smart-light and SNIP Washing-machine datasets \cite{coucke2019efficient}, and intent classification (IC) was tested on the FSC dataset \citep{lugosch2019speech}. Accuracy (ACC) served as the metric for both KE and IC tasks. To evaluate the model's QA capabilities, we utilized TTS\footnote{https://www.xfyun.cn/service/offline\_tts} to convert the Webglm-QA \cite{liu2023webglm} dataset and the BoolQA \cite{clark2019boolq} dataset into a speech format. For the WebQA dataset, BertScore \cite{zhang2019bertscore} was used as the evaluation metric, and for BoolQA, ACC was used as the evaluation metric. All the instructions used for evaluation are present in Appendix~\ref{appendix:three}. 

\subsection{Baseline}
We conducted a comparative analysis of the self-powered LSM, benchmarking it against baseline models trained on vanilla ASR speech instructional data as well as previous methodologies, Qwen-Audio \cite{chu2023qwen} and BLSP \cite{wang2023blsp}. The core idea of BLSP, which involves aligning model behavior using a continuation task, can be seen as an instance of our method. To ensure comparability, we replicated it using identical datasets and architectural frameworks. All the models we evaluated share the same instructions.

\begin{table*}[tbp]
\centering
\renewcommand{\arraystretch}{1.1}
\scalebox{0.77}{
\begin{tabular}{lcccccccccc}
\toprule
\multirow{2}{*}{\bf Method} & \multicolumn{2}{c}{\bf ASR$\downarrow$} &  \multicolumn{2}{c}{\bf ST$\uparrow$} &\bf ER$\uparrow$ &\multicolumn{2}{c}{\bf KE$\uparrow$} & \bf IC$\uparrow$ & \multicolumn{2}{c}{\bf QA$\uparrow$}   \\
\cmidrule(lr){2-3} \cmidrule(lr){4-5}  \cmidrule(lr){6-6}  \cmidrule(lr){7-8}  \cmidrule(lr){9-9} \cmidrule(lr){10-11}
& Clean & Other  & CoVoST & MuST-C & MELD & Light & Water & FSC & WebQ & BoolQ  \\
\midrule
Emotion $+$ Intent &  100.0 & 100.0  &  0.0 &  0.0  & \bf 71.7 &  0.0 &  0.0 & \bf 99.6 & 43.2 & 0.0  \\
\hdashline
Translation  & 100.0 & 100.0 & 4.5  & 0.0 & 0.0 & 0.0 & 0.0 & 0.0 & 42.2 & 0.0    \\
~~w/ Self-Powered Aug.  & 90.8 &  94.1 &  10.2  & 6.2 & 29.4 & 15.5 & 15.1 & 2.0 & 50.7 & 48.6 \\
\hdashline
Self-Powered LSM & \bf 3.8 & \bf 8.0  &\bf 17.5 & \bf 21.4 & 51.4 & \bf 30.7 & \bf 34.9 & 47.2 & \bf 72.4 & \bf 59.8  \\
\bottomrule
\end{tabular}}
\caption{Comparison of the LSM training on diverse ground-truth and self-powered datasets.}
\label{tab:groud}
\end{table*}

\subsection{Main Result}
Our results, as presented in Table \ref{tab:main_all}, substantiate that our methodology effectively enhances the speech modality capabilities of LLMs for LSMs.

\paragraph{Compared with Vanilla IT }
Initially, we evaluated a baseline model trained on vanilla speech instructional data (Model ``3''), noting substantial overfitting in ASR tasks. 
This finding supports our hypothesis in Section~\ref{understand} that the direct fine-tuning of LSMs induces significant speech anchor bias, leading to modality imbalance when LLMs are employed to augment the speech capabilities of LSMs.
In contrast, training with self-powered data (Model ``5'') significantly enhances ASR performance and bolsters the model’s ability to execute instructions across a variety of speech-driven tasks. These findings confirm that our proposed method efficiently imparts speech comprehension and instruction-following abilities to the LSMs.

\paragraph{Compared with Prior Related Studies} 
Compared to Qwen-Audio (Model ``1''), which trained using ground-truth data including ASR and CoVoST ST datasets, our method—employing a limited subset of ASR data and no ground-truth translation data—demonstrated superior performance across all SLU and QA tasks. Although this resulted in slightly lower in-domain performance, self-powered LSM excelled in ST tasks on the out-of-domain MuST-C dataset, showcasing its enhanced generalization capabilities derived from training on generated pseudo-data.
Compared to models like BLSP (Model ``4''), our method exhibited notable improvements in all assessed tasks, particularly in ASR. This underscores our method's superiority over using continuation instructions to augment the dataset, highlighting the self-powered augmentation method's comprehensive enhancement of model training and task performance.

\paragraph{Compared across Different Speech Encoders}
To further investigate the impact of varying speech encoder sizes, we employed self-powered LSM across three whisper encoder scales: small, medium, and large-v2.
The results (Model ``5'', Model ``6'' and Model ``7'') demonstrate that employing more powerful speech encoders enhances LSM performance. With the implementation of whisper-large, self-powered LSM achieves the superior results especially on SLU and QA tasks.

\section{Analysis}

\subsection{Ablation Study on Self-Powered LSM}
\paragraph{Impact of Different Training Strategy} Table~\ref{tab:strategy} details the performance of self-powered LSM under various training strategies. Omitting LLM fine-tuning significantly reduces performance across speech-related tasks, particularly impacting QA capabilities, thus underscoring the crucial role of LLM fine-tuning in achieving instruction alignment.
Improvements were observed in SLU and QA tasks when employing LoRA for LLM. However, performance declined notably in ASR and ST tasks. As discussed by \citet{tang2023salmonn}, training the LSM model using fine-tuned LoRA involves intricate design and adjustments of LoRA rank. In contrast, our training method not only simplifies training but also achieves superior performance, demonstrating that full fine-tuning remains the simplest and most superior strategy for expanding LLM with speech capabilities.

\begin{table*}[htbp]
\centering
\renewcommand{\arraystretch}{1.1}
\scalebox{0.77}{
\begin{tabular}{lcccccccccc}
\toprule
\multirow{2}{*}{\bf ID~~~Method} & \multicolumn{2}{c}{\bf ASR$\downarrow$} &  \multicolumn{2}{c}{\bf ST$\uparrow$} &\bf ER$\uparrow$ &\multicolumn{2}{c}{\bf KE$\uparrow$} & \bf IC$\uparrow$ & \multicolumn{2}{c}{\bf QA$\uparrow$}   \\
\cmidrule(lr){2-3} \cmidrule(lr){4-5}  \cmidrule(lr){6-6}  \cmidrule(lr){7-8}  \cmidrule(lr){9-9} \cmidrule(lr){10-11}
& Clean & Other  & CoVoST & MuST-C & MELD & Light & Water & FSC & WebQ & BoolQ  \\
\midrule
Vanilla IT  & \bf 40.0 & \bf 42.6 &  0.0 & 0.0 & 0.2 & 0.0 & 0.0 & 0.0 &42.2 & 0.0  \\

Self-Powered Llama3  & 50.9 & 49.1 & \bf 15.3 &  \bf 14.1 & \bf 51.4 & \bf 56.5 & \bf 35.8 & \bf 86.8 & \bf 67.0 & \bf 60.6 \\
\bottomrule
\end{tabular}}
\caption{Results of the Llama3-8B-Instruct as the backbone LLM.}
\label{tab:llama}
\end{table*}

\begin{figure}[tbp]
    \centering
    \includegraphics[scale=0.17]{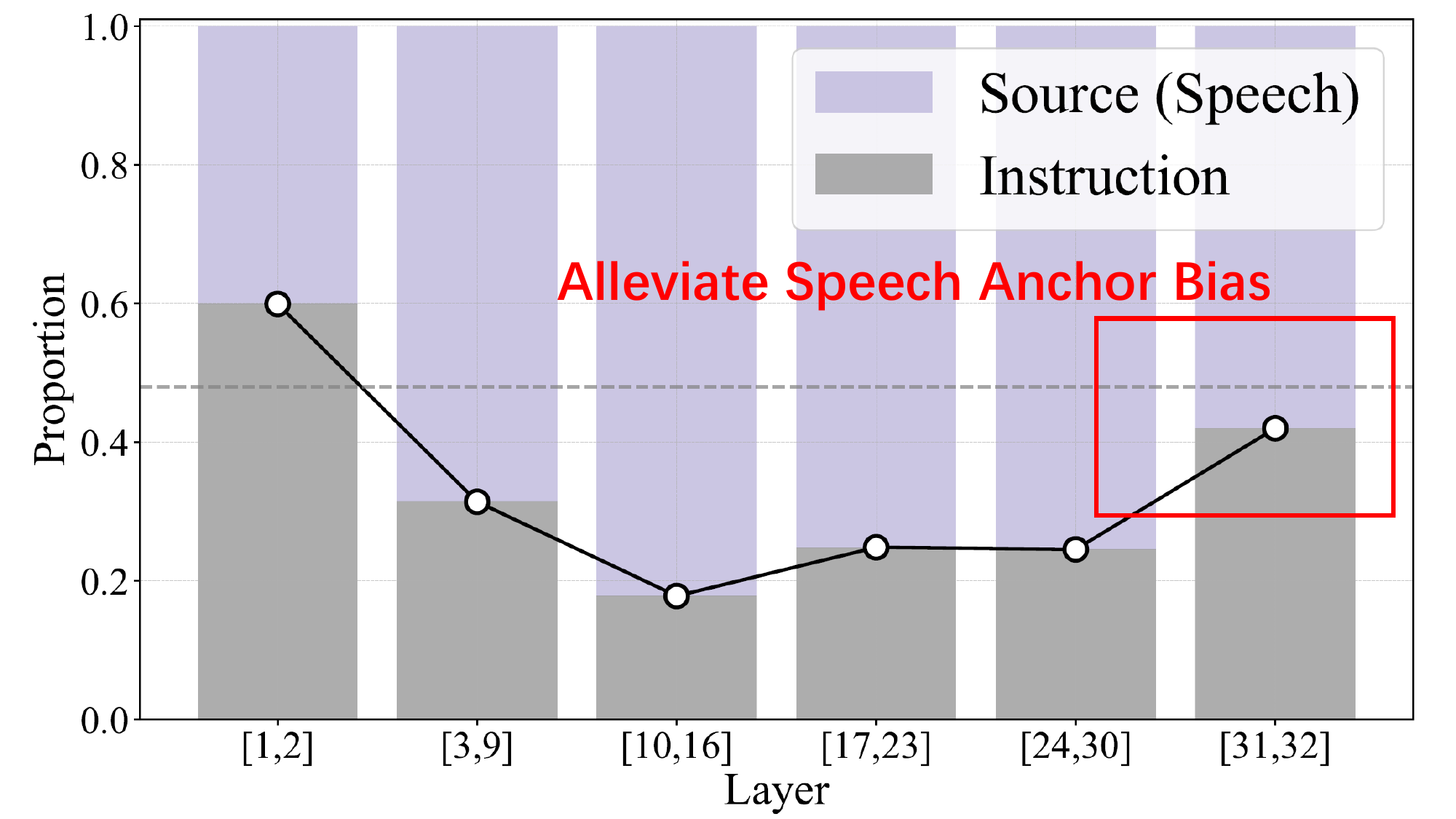}
    \caption{Layer-wise behaviors in self-powered LSM.}
    \label{fig:compare}
\end{figure}

\paragraph{Impact of Self-Powered Data}
To assess the efficacy of self-powered data in augmenting the speech modal capabilities of LLMs, we conducted experiments using various ground-truth and self-powered datasets. As indicated in Table~\ref{tab:groud}, the LSM, when fine-tuned with ground-truth datasets for specific SLU tasks (``Emotion $+$ Intent'') and the CoVoST En-Zh training set (``Translation''), exhibited proficiency only in in-domain tasks. This limitation suggested a lack of instruction-following capabilities, as the model failed to generalize beyond its training conditions. In contrast, training the LSM with self-powered En-Zh translation data of equivalent size endowed the model with nascent instruction generalization abilities. Further extensive training exclusively with fully self-powered data significantly enhanced performance across all tasks.

\begin{figure}[tbp]
    \centering
    \includegraphics[width=0.43\textwidth, height=0.13\textheight]{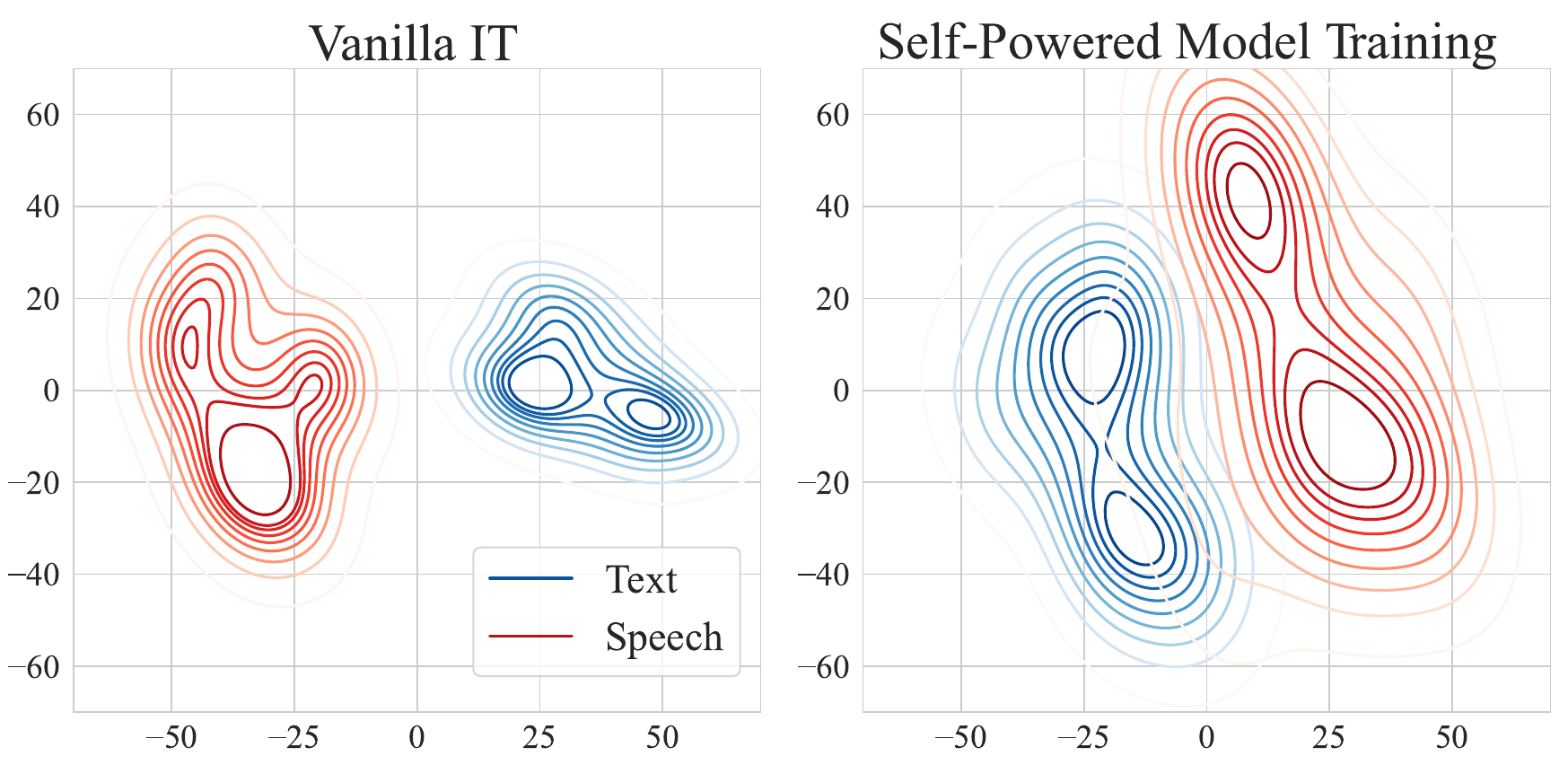}
    \caption{T-SNE visualization of representations from LSMs trained with vanilla IT and self-powered methods. }
    \label{fig:visual}
\end{figure}
\subsection{Effectiveness of Self-Powered LSM}
\paragraph{Mitigating Speech Anchor Bias}
In Section~\ref{sec:4.2}, we posit that using self-powered data can enhance the salience of instructional inputs in LSM training. To substantiate this hypothesis, we conducted a layer-wise behavioral analysis of self-powered LSM. Figure~\ref{fig:compare} shows that utilizing augmentation data in LSM training markedly improves the model's sensitivity to instructional cues, especially in the terminal layers (where instruction proportion rises from 0.2 to 0.4). This pattern aligns closely with that observed in instruction-following LLMs, Alpaca and Vicuna, showing that its efficacy in mitigating the speech anchor bias on LSM.

\paragraph{Boosting Speech-Text Alignment} 
We developed a specialized test set of 1,200 examples using five instructional templates (detailed in Appendix~\ref{appendix:three}) to explore representation alignment between speech and text with the vanilla IT approach and the self-powered LSM. The models processed speech or transcription inputs following identical instructions and extracted averaged representations from the hidden states of the final layer.
We applied bivariate kernel density estimation \cite{parzen1962estimation} and utilized the T-SNE technique to reduce data dimensions to a two-dimensional space \cite{van2008visualizing}.
The findings, depicted in Figure~\ref{fig:visual}, indicate that the self-powered LSM approach significantly improves the proximity between the representations of paired speech and text inputs compared to the vanilla IT method, thereby demonstrating the effectiveness of our methodology in modality fusion. 

\begin{table}[t]
\centering
\scalebox{0.78}{
\begin{tabular}{l cc cc c}
\toprule
\bf Model &  \bf STEM & \bf Humanities & \bf Society & \bf Other & \bf Avg. \\
\midrule
LLM &  \bf 39.5 & 45.7& \bf 58.1 & \bf 57.4& \bf 49.8\\ 
LSM  & 38.7 & \bf 45.9 &57.8 &56.4 & 49.4\\ 

\bottomrule
\end{tabular}}
\caption{Text-only results on MMLU benchmark. ``LSM'' denotes our Self-Powered LSM and ``LLM'' refers to its backbone LLM, Vicuna-1.5-7B.}
\label{tab:textresult}
\end{table}

\paragraph{Maintaining General Textual Performance}
To explore how the text-only performance of LSM has been affected as a result of the fine-tuning of LLM, we supplemented this with a comparison of the self-powered LSM against Vicuna-7B 1.5 on the MMLU benchmark. We conducted tests using an open-instruction \cite{wang2023far} script on the MMLU test set. The experimental results, as shown in Table~\ref{tab:textresult}, demonstrate that even after training with self-powered speech data, the model achieves comparable results on MMLU, indicating that the LLM’s general capabilities are not compromised and signaling that there is no issue of catastrophic forgetting. This shows that our method can efficiently expand the model's speech capabilities without adversely affecting its original textual question-answering proficiency. 
\subsection{Various Model Adapting}
To assess the generalizability of our approach, we utilized the Llama3-8B-Instruct as the backbone LLM and implemented self-powered data generated by Llama3-Instruct for training. As shown in Table~\ref{tab:llama}, our method consistently delivers high performance across ST, SLU, and QA tasks. This underscores the effectiveness and adaptability of our self-powered method. Furthermore, we find that both vanilla ASR data and self-powered data contribute minimally to enhancing ASR performance.
Although Llama3-Instruct has enhanced capabilities, it often produces additional content and detail, leading to mismatches with the WER evaluation metric in ASR scenarios. We leave this issue for further research to explore deeper evaluations.

\begin{table}[t]
\centering
\scalebox{0.7}{
\begin{tabular}{l cc cc cc}
\toprule
\multirow{2}{*}{\bf Method} &  \multicolumn{2}{c}{\bf ST} &\multicolumn{2}{c}{\bf SLU} &\multicolumn{2}{c}{\bf QA} \\
\cmidrule(lr){2-3} \cmidrule(lr){4-5} \cmidrule(lr){6-7} 
 &  \bf{Result} & \bf{Speed} & \bf{Result} & \bf{Speed} & \bf{Result} & \bf{Speed} \\ 
\midrule
Cascade &  \bf 23.1 & 1.0$\times$& 33.2&$1.0\times$ &\bf 68.9 &$1.0\times$\\ 
End-to-End   & 22.3 &\bf 3.8$\times$ &\bf 49.0 &\bf 3.0$\times$ & 67.5
 &\bf 3.1$\times$\\ 
\bottomrule
\end{tabular}}
\caption{Comparison of performance and speed between our end-to-end LSM and a cascaded approach that uses Whisper for speech transcription, followed by Vicuna-1.5-7B for responding to instructions.}
\label{tab:speed}
\end{table}

\subsection{Cascade vs. End-to-End LSMs}
We compare our end-to-end LSM with a cascade pipeline. The cascade pipeline employs a two-step approach, initially utilizing Whisper to transcribe speech into text, then leveraging Vicuna-1.5-7B to respond to instructions. Table~\ref{tab:speed} reveals that although the cascade method can achieve slightly better results in ST and QA, its performance in SLU is limited. In contrast, our end-to-end LSM delivers decent performance on all tasks, with superior inference speed.
This distinction highlights the efficiency and potential application benefits of the end-to-end approach. We aim to clarify that the goal of this research is not to outperform the pipeline method but rather to explore the potential of end-to-end models and offer new insights, paving the way for future advancements.

\section{Conclusion}
In this study, we aim to expand the capabilities of LLMs in speech modalities by exclusively utilizing ASR data. To achieve this, we explore the instruction-following dynamics within LSMs, identifying a significant issue that we term \textit{speech anchor bias}. This bias highlights LSMs' tendency to unduly emphasize speech inputs. To rectify this, we developed a novel self-powered method that leverages augmented ASR data generated by the models to enhance LSMs' instruction-following capabilities. Our experiments across multiple speech tasks demonstrate that the self-powered LSM significantly mitigates the speech anchor bias, substantially improving LSMs' fidelity to instructions.

\section*{Limitations}
We acknowledge several limitations in our approach to understanding and enhancing large speech models: (1) We employ a norm-based attention metric to comprehensively consider the attention mechanism and multimodal input features, aiming to explore the intrinsic information flow mechanisms within LSM. However, this numerical interpretation is often superficial. In the future, we plan to analyze from deeper and more diverse perspectives \cite{yu-etal-2023-promptst} to further investigate the causes of overfitting during the training of LSM. (2) While we aim for a comprehensive evaluation, our assessments are limited to specific tasks and may lack robustness due to the metrics and datasets used, highlighting the need for further research under broader conditions. (3) Although our method enhances the end-to-end speech capabilities of LSMs, a performance gap remains compared to the cascaded approach, particularly in ST tasks. However, we believe our method provides valuable insights and encourages the development of end-to-end multimodal large models. (4) Concurrent research indicates that carefully curated, smaller, high-quality datasets can significantly enhance a model’s ability to follow instructions \citep{zhou2023lima,li2023textbooks,xu2023paradigm,liu2024selectit}. Despite its size, our dataset offers significant opportunities for refinement. We will make all data publicly available to promote ongoing academic exploration.

\section*{Ethics Statement}
In conducting our research, we strictly adhere to all applicable laws and regulations. Additionally, we take ethical considerations seriously and adhere to the standards of the ACL Ethics Policy. This paper focuses on understanding and improving large speech model. To support reproducibility, both the datasets and code used in this study are available to other researchers upon request. We have made every effort to present our findings and conclusions in an objective and accurate manner.

\section*{Acknowledgments}
This work was supported in part by the National Natural Science Foundation of China (Grant No. 62206076), Guangdong Basic and Applied Basic Research Foundation (Grant No. 2024A1515011491), Shenzhen Science and Technology Program (Grant Nos. ZDSYS20230626091203008, KJZD20231023094700001, RCBS20221008093121053), Shenzhen College Stability Support Plan (Grant Nos. GXWD20220811173340003, GXWD20220817123150002). Dacheng Tao is supported by the National Research Foundation, Singapore, and the CyberSG R$\&$D Programme Office (“CRPO”), under the National Cybersecurity R$\&$D Programme (“NCRP”), RIE2025 NCRP Funding Initiative (Award CRPO-GC1-NTU-002). We would like to thank the anonymous reviewers and meta-reviewer for their insightful suggestions.

\bibliography{acl_latex}

\begin{thebibliography}{61}
\expandafter\ifx\csname natexlab\endcsname\relax\def\natexlab#1{#1}\fi

\bibitem[{Anil et~al.(2023)Anil, Dai, Firat, Johnson, Lepikhin, Passos, Shakeri, Taropa, Bailey, Chen et~al.}]{palm2}
Rohan Anil, Andrew~M Dai, Orhan Firat, Melvin Johnson, Dmitry Lepikhin, Alexandre Passos, Siamak Shakeri, Emanuel Taropa, Paige Bailey, Zhifeng Chen, et~al. 2023.
\newblock \href {https://arxiv.org/abs/2305.10403} {{PaLM} 2 technical report}.
\newblock \emph{ArXiv preprint}, abs/2305.10403.

\bibitem[{Ardila et~al.(2020)Ardila, Branson, Davis, Kohler, Meyer, Henretty, Morais, Saunders, Tyers, and Weber}]{ardila2019common}
Rosana Ardila, Megan Branson, Kelly Davis, Michael Kohler, Josh Meyer, Michael Henretty, Reuben Morais, Lindsay Saunders, Francis Tyers, and Gregor Weber. 2020.
\newblock \href {https://aclanthology.org/2020.lrec-1.520} {Common voice: A massively-multilingual speech corpus}.
\newblock In \emph{Proceedings of the Twelfth Language Resources and Evaluation Conference}, pages 4218--4222, Marseille, France. European Language Resources Association.

\bibitem[{Bibal et~al.(2022)Bibal, Cardon, Alfter, Wilkens, Wang, Fran{\c{c}}ois, and Watrin}]{bibal2022attention}
Adrien Bibal, R{\'e}mi Cardon, David Alfter, Rodrigo Wilkens, Xiaoou Wang, Thomas Fran{\c{c}}ois, and Patrick Watrin. 2022.
\newblock \href {https://doi.org/10.18653/v1/2022.acl-long.269} {Is attention explanation? an introduction to the debate}.
\newblock In \emph{Proceedings of the 60th Annual Meeting of the Association for Computational Linguistics (Volume 1: Long Papers)}, pages 3889--3900, Dublin, Ireland. Association for Computational Linguistics.

\bibitem[{Brown et~al.(2020)Brown, Mann, Ryder, Subbiah, Kaplan, Dhariwal, Neelakantan, Shyam, Sastry, Askell, Agarwal, Herbert{-}Voss, Krueger, Henighan, Child, Ramesh, Ziegler, Wu, Winter, Hesse, Chen, Sigler, Litwin, Gray, Chess, Clark, Berner, McCandlish, Radford, Sutskever, and Amodei}]{gpt3}
Tom~B. Brown, Benjamin Mann, Nick Ryder, Melanie Subbiah, Jared Kaplan, Prafulla Dhariwal, Arvind Neelakantan, Pranav Shyam, Girish Sastry, Amanda Askell, Sandhini Agarwal, Ariel Herbert{-}Voss, Gretchen Krueger, Tom Henighan, Rewon Child, Aditya Ramesh, Daniel~M. Ziegler, Jeffrey Wu, Clemens Winter, Christopher Hesse, Mark Chen, Eric Sigler, Mateusz Litwin, Scott Gray, Benjamin Chess, Jack Clark, Christopher Berner, Sam McCandlish, Alec Radford, Ilya Sutskever, and Dario Amodei. 2020.
\newblock \href {https://proceedings.neurips.cc/paper/2020/hash/1457c0d6bfcb4967418bfb8ac142f64a-Abstract.html} {Language models are few-shot learners}.
\newblock In \emph{Advances in Neural Information Processing Systems 33: Annual Conference on Neural Information Processing Systems 2020, NeurIPS 2020, December 6-12, 2020, virtual}.

\bibitem[{Chen et~al.(2023{\natexlab{a}})Chen, Han, Zhao, Zhang, Shi, Xu, and Xu}]{chen2023xllm}
Feilong Chen, Minglun Han, Haozhi Zhao, Qingyang Zhang, Jing Shi, Shuang Xu, and Bo~Xu. 2023{\natexlab{a}}.
\newblock \href {https://arxiv.org/abs/2305.04160} {{X-LLM}: {B}ootstrapping advanced large language models by treating multi-modalities as foreign languages}.
\newblock \emph{ArXiv preprint}, abs/2305.04160.

\bibitem[{Chen et~al.(2021)Chen, Chai, Wang, Du, Zhang, Weng, Su, Povey, Trmal, Zhang, Jin, Khudanpur, Watanabe, Zhao, Zou, Li, Yao, Wang, You, and Yan}]{chen2021gigaspeech}
Guoguo Chen, Shuzhou Chai, Guan{-}Bo Wang, Jiayu Du, Wei{-}Qiang Zhang, Chao Weng, Dan Su, Daniel Povey, Jan Trmal, Junbo Zhang, Mingjie Jin, Sanjeev Khudanpur, Shinji Watanabe, Shuaijiang Zhao, Wei Zou, Xiangang Li, Xuchen Yao, Yongqing Wang, Zhao You, and Zhiyong Yan. 2021.
\newblock \href {https://doi.org/10.21437/Interspeech.2021-1965} {Gigaspeech: An evolving, multi-domain {ASR} corpus with 10, 000 hours of transcribed audio}.
\newblock In \emph{Interspeech 2021, 22nd Annual Conference of the International Speech Communication Association, Brno, Czechia, 30 August - 3 September 2021}, pages 3670--3674. {ISCA}.

\bibitem[{Chen et~al.(2023{\natexlab{b}})Chen, Chu, Gao, Li, Hu, Zhou, Xu, Ma, Wang, Zheng et~al.}]{chen2023lauragpt}
Qian Chen, Yunfei Chu, Zhifu Gao, Zerui Li, Kai Hu, Xiaohuan Zhou, Jin Xu, Ziyang Ma, Wen Wang, Siqi Zheng, et~al. 2023{\natexlab{b}}.
\newblock \href {https://arxiv.org/abs/2310.04673} {Lauragpt: Listen, attend, understand, and regenerate audio with gpt}.
\newblock \emph{ArXiv preprint}, abs/2310.04673.

\bibitem[{Chiang et~al.(2023)Chiang, Li, Lin, Sheng, Wu, Zhang, Zheng, Zhuang, Zhuang, Gonzalez et~al.}]{chiang2023vicuna}
Wei-Lin Chiang, Zhuohan Li, Zi~Lin, Ying Sheng, Zhanghao Wu, Hao Zhang, Lianmin Zheng, Siyuan Zhuang, Yonghao Zhuang, Joseph~E Gonzalez, et~al. 2023.
\newblock \href {https://lmsys. org/blog/2023-03-30-vicuna} {Vicuna: An open-source chatbot impressing gpt-4 with 90\%* chatgpt quality}.
\newblock \emph{See https://vicuna. lmsys. org (accessed 14 April 2023)}.

\bibitem[{Chu et~al.(2023)Chu, Xu, Zhou, Yang, Zhang, Yan, Zhou, and Zhou}]{chu2023qwen}
Yunfei Chu, Jin Xu, Xiaohuan Zhou, Qian Yang, Shiliang Zhang, Zhijie Yan, Chang Zhou, and Jingren Zhou. 2023.
\newblock \href {https://arxiv.org/abs/2311.07919} {Qwen-audio: Advancing universal audio understanding via unified large-scale audio-language models}.
\newblock \emph{arXiv preprint arXiv:2311.07919}.

\bibitem[{Clark et~al.(2019{\natexlab{a}})Clark, Lee, Chang, Kwiatkowski, Collins, and Toutanova}]{clark2019boolq}
Christopher Clark, Kenton Lee, Ming-Wei Chang, Tom Kwiatkowski, Michael Collins, and Kristina Toutanova. 2019{\natexlab{a}}.
\newblock \href {https://arxiv.org/abs/1905.10044} {Boolq: Exploring the surprising difficulty of natural yes/no questions}.
\newblock In \emph{Proceedings of the 2019 Conference of the North American Chapter of the Association for Computational Linguistics: Human Language Technologies, Volume 1 (Long and Short Papers)}, pages 2924--2936.

\bibitem[{Clark et~al.(2019{\natexlab{b}})Clark, Khandelwal, Levy, and Manning}]{clark19}
Kevin Clark, Urvashi Khandelwal, Omer Levy, and Christopher~D. Manning. 2019{\natexlab{b}}.
\newblock \href {https://doi.org/10.18653/v1/W19-4828} {What does {BERT} look at? an analysis of {BERT}{'}s attention}.
\newblock In \emph{Proceedings of the 2019 ACL Workshop BlackboxNLP: Analyzing and Interpreting Neural Networks for NLP}, pages 276--286, Florence, Italy. Association for Computational Linguistics.

\bibitem[{Coucke et~al.(2019)Coucke, Chlieh, Gisselbrecht, Leroy, Poumeyrol, and Lavril}]{coucke2019efficient}
Alice Coucke, Mohammed Chlieh, Thibault Gisselbrecht, David Leroy, Mathieu Poumeyrol, and Thibaut Lavril. 2019.
\newblock \href {https://doi.org/10.1109/ICASSP.2019.8683474} {Efficient keyword spotting using dilated convolutions and gating}.
\newblock In \emph{{IEEE} International Conference on Acoustics, Speech and Signal Processing, {ICASSP} 2019, Brighton, United Kingdom, May 12-17, 2019}, pages 6351--6355. {IEEE}.

\bibitem[{Di~Gangi et~al.(2019)Di~Gangi, Cattoni, Bentivogli, Negri, and Turchi}]{di2019must}
Mattia~A. Di~Gangi, Roldano Cattoni, Luisa Bentivogli, Matteo Negri, and Marco Turchi. 2019.
\newblock \href {https://doi.org/10.18653/v1/N19-1202} {{M}u{ST}-{C}: a {M}ultilingual {S}peech {T}ranslation {C}orpus}.
\newblock In \emph{Proceedings of the 2019 Conference of the North {A}merican Chapter of the Association for Computational Linguistics: Human Language Technologies, Volume 1 (Long and Short Papers)}, pages 2012--2017, Minneapolis, Minnesota. Association for Computational Linguistics.

\bibitem[{Hu et~al.(2022)Hu, Shen, Wallis, Allen{-}Zhu, Li, Wang, Wang, and Chen}]{hu2021lora}
Edward~J. Hu, Yelong Shen, Phillip Wallis, Zeyuan Allen{-}Zhu, Yuanzhi Li, Shean Wang, Lu~Wang, and Weizhu Chen. 2022.
\newblock \href {https://openreview.net/forum?id=nZeVKeeFYf9} {Lora: Low-rank adaptation of large language models}.
\newblock In \emph{The Tenth International Conference on Learning Representations, {ICLR} 2022, Virtual Event, April 25-29, 2022}. OpenReview.net.

\bibitem[{Huang et~al.(2024)Huang, Li, Yang, Shi, Chang, Ye, Wu, Hong, Huang, Liu et~al.}]{huang2023udiogpt}
Rongjie Huang, Mingze Li, Dongchao Yang, Jiatong Shi, Xuankai Chang, Zhenhui Ye, Yuning Wu, Zhiqing Hong, Jiawei Huang, Jinglin Liu, et~al. 2024.
\newblock \href {https://ojs.aaai.org/index.php/AAAI/article/view/30570} {Audiogpt: Understanding and generating speech, music, sound, and talking head}.
\newblock In \emph{Proceedings of the AAAI Conference on Artificial Intelligence}, volume~38, pages 23802--23804.

\bibitem[{Kobayashi et~al.(2020)Kobayashi, Kuribayashi, Yokoi, and Inui}]{kobayashi2020attention}
Goro Kobayashi, Tatsuki Kuribayashi, Sho Yokoi, and Kentaro Inui. 2020.
\newblock \href {https://doi.org/10.18653/v1/2020.emnlp-main.574} {Attention is not only a weight: Analyzing transformers with vector norms}.
\newblock In \emph{Proceedings of the 2020 Conference on Empirical Methods in Natural Language Processing (EMNLP)}, pages 7057--7075, Online. Association for Computational Linguistics.

\bibitem[{Lakhotia et~al.(2021)Lakhotia, Kharitonov, Hsu, Adi, Polyak, Bolte, Nguyen, Copet, Baevski, Mohamed, and Dupoux}]{lakhotia2021generative}
Kushal Lakhotia, Eugene Kharitonov, Wei-Ning Hsu, Yossi Adi, Adam Polyak, Benjamin Bolte, Tu-Anh Nguyen, Jade Copet, Alexei Baevski, Abdelrahman Mohamed, and Emmanuel Dupoux. 2021.
\newblock \href {https://doi.org/10.1162/tacl_a_00430} {On generative spoken language modeling from raw audio}.
\newblock \emph{Transactions of the Association for Computational Linguistics}, 9:1336--1354.

\bibitem[{Li et~al.(2023{\natexlab{a}})Li, Li, Savarese, and Hoi}]{li2023blip}
Junnan Li, Dongxu Li, Silvio Savarese, and Steven Hoi. 2023{\natexlab{a}}.
\newblock \href {https://arxiv.org/abs/2301.12597} {Blip-2: Bootstrapping language-image pre-training with frozen image encoders and large language models}.
\newblock \emph{ArXiv preprint}, abs/2301.12597.

\bibitem[{Li et~al.(2023{\natexlab{b}})Li, Bubeck, Eldan, Del~Giorno, Gunasekar, and Lee}]{li2023textbooks}
Yuanzhi Li, S{\'e}bastien Bubeck, Ronen Eldan, Allie Del~Giorno, Suriya Gunasekar, and Yin~Tat Lee. 2023{\natexlab{b}}.
\newblock \href {https://arxiv.org/abs/2309.05463} {Textbooks are all you need ii: phi-1.5 technical report}.
\newblock \emph{ArXiv preprint}, abs/2309.05463.

\bibitem[{Liu et~al.(2023{\natexlab{a}})Liu, Li, Wu, and Lee}]{liu2023visual}
Haotian Liu, Chunyuan Li, Qingyang Wu, and Yong~Jae Lee. 2023{\natexlab{a}}.
\newblock \href {https://arxiv.org/abs/2304.08485} {Visual instruction tuning}.
\newblock \emph{ArXiv preprint}, abs/2304.08485.

\bibitem[{Liu et~al.(2024)Liu, Liu, Wong, Li, Wang, Hu, and Zhang}]{liu2024selectit}
Liangxin Liu, Xuebo Liu, Derek~F Wong, Dongfang Li, Ziyi Wang, Baotian Hu, and Min Zhang. 2024.
\newblock \href {https://arxiv.org/abs/2402.16705} {Selectit: Selective instruction tuning for large language models via uncertainty-aware self-reflection}.
\newblock \emph{arXiv preprint arXiv:2402.16705}.

\bibitem[{Liu et~al.(2023{\natexlab{b}})Liu, Lai, Yu, Xu, Zeng, Du, Zhang, Dong, and Tang}]{liu2023webglm}
Xiao Liu, Hanyu Lai, Hao Yu, Yifan Xu, Aohan Zeng, Zhengxiao Du, Peng Zhang, Yuxiao Dong, and Jie Tang. 2023{\natexlab{b}}.
\newblock \href {https://dl.acm.org/doi/abs/10.1145/3580305.3599931} {Webglm: Towards an efficient web-enhanced question answering system with human preferences}.
\newblock In \emph{Proceedings of the 29th ACM SIGKDD Conference on Knowledge Discovery and Data Mining}, pages 4549--4560.

\bibitem[{Liu et~al.(2021)Liu, Wang, Wong, Ding, Chao, and Tu}]{liu2020understanding}
Xuebo Liu, Longyue Wang, Derek~F. Wong, Liang Ding, Lidia~S. Chao, and Zhaopeng Tu. 2021.
\newblock \href {https://openreview.net/forum?id=n1HD8M6WGn} {Understanding and improving encoder layer fusion in sequence-to-sequence learning}.
\newblock In \emph{9th International Conference on Learning Representations, {ICLR} 2021, Virtual Event, Austria, May 3-7, 2021}. OpenReview.net.

\bibitem[{Lugosch et~al.(2019)Lugosch, Ravanelli, Ignoto, Tomar, and Bengio}]{lugosch2019speech}
Loren Lugosch, Mirco Ravanelli, Patrick Ignoto, Vikrant~Singh Tomar, and Yoshua Bengio. 2019.
\newblock \href {https://doi.org/10.21437/Interspeech.2019-2396} {Speech model pre-training for end-to-end spoken language understanding}.
\newblock In \emph{Interspeech 2019, 20th Annual Conference of the International Speech Communication Association, Graz, Austria, 15-19 September 2019}, pages 814--818. {ISCA}.

\bibitem[{Madaan et~al.(2024)Madaan, Tandon, Gupta, Hallinan, Gao, Wiegreffe, Alon, Dziri, Prabhumoye, Yang et~al.}]{madaan2024self}
Aman Madaan, Niket Tandon, Prakhar Gupta, Skyler Hallinan, Luyu Gao, Sarah Wiegreffe, Uri Alon, Nouha Dziri, Shrimai Prabhumoye, Yiming Yang, et~al. 2024.
\newblock \href {https://proceedings.neurips.cc/paper_files/paper/2023/hash/91edff07232fb1b55a505a9e9f6c0ff3-Abstract-Conference.html} {Self-refine: Iterative refinement with self-feedback}.
\newblock \emph{Advances in Neural Information Processing Systems}, 36.

\bibitem[{Mohamed et~al.(2022)Mohamed, Lee, Borgholt, Havtorn, Edin, Igel, Kirchhoff, Li, Livescu, Maal{\o}e et~al.}]{mohamed2022self}
Abdelrahman Mohamed, Hung-yi Lee, Lasse Borgholt, Jakob~D Havtorn, Joakim Edin, Christian Igel, Katrin Kirchhoff, Shang-Wen Li, Karen Livescu, Lars Maal{\o}e, et~al. 2022.
\newblock \href {https://arxiv.org/abs/2205.10643} {Self-supervised speech representation learning: A review}.
\newblock \emph{ArXiv preprint}, abs/2205.10643.

\bibitem[{OpenAI(2023)}]{gpt4}
OpenAI. 2023.
\newblock \href {https://arxiv.org/abs/2303.08774} {{GPT-4} technical report}.
\newblock \emph{ArXiv preprint}, abs/2303.08774.

\bibitem[{Pan et~al.(2023)Pan, Wu, Gaur, Sivasankaran, Chen, Liu, and Li}]{pan2023cosmic}
Jing Pan, Jian Wu, Yashesh Gaur, Sunit Sivasankaran, Zhuo Chen, Shujie Liu, and Jinyu Li. 2023.
\newblock \href {https://arxiv.org/abs/2311.02248} {Cosmic: Data efficient instruction-tuning for speech in-context learning}.
\newblock \emph{arXiv e-prints}, pages arXiv--2311.

\bibitem[{Panayotov et~al.(2015)Panayotov, Chen, Povey, and Khudanpur}]{panayotov2015librispeech}
Vassil Panayotov, Guoguo Chen, Daniel Povey, and Sanjeev Khudanpur. 2015.
\newblock \href {https://doi.org/10.1109/ICASSP.2015.7178964} {Librispeech: An {ASR} corpus based on public domain audio books}.
\newblock In \emph{2015 {IEEE} International Conference on Acoustics, Speech and Signal Processing, {ICASSP} 2015, South Brisbane, Queensland, Australia, April 19-24, 2015}, pages 5206--5210. {IEEE}.

\bibitem[{Parzen(1962)}]{parzen1962estimation}
Emanuel Parzen. 1962.
\newblock \href {https://www.jstor.org/stable/2237880} {On estimation of a probability density function and mode}.
\newblock \emph{The annals of mathematical statistics}, 33(3):1065--1076.

\bibitem[{Popuri et~al.(2022)Popuri, Chen, Wang, Pino, Adi, Gu, Hsu, and Lee}]{popuri2022enhanced}
Sravya Popuri, Peng-Jen Chen, Changhan Wang, Juan Pino, Yossi Adi, Jiatao Gu, Wei-Ning Hsu, and Ann Lee. 2022.
\newblock \href {https://arxiv.org/abs/2204.02967} {Enhanced direct speech-to-speech translation using self-supervised pre-training and data augmentation}.
\newblock \emph{ArXiv preprint}, abs/2204.02967.

\bibitem[{Poria et~al.(2019)Poria, Hazarika, Majumder, Naik, Cambria, and Mihalcea}]{poria2018meld}
Soujanya Poria, Devamanyu Hazarika, Navonil Majumder, Gautam Naik, Erik Cambria, and Rada Mihalcea. 2019.
\newblock \href {https://doi.org/10.18653/v1/P19-1050} {{MELD}: A multimodal multi-party dataset for emotion recognition in conversations}.
\newblock In \emph{Proceedings of the 57th Annual Meeting of the Association for Computational Linguistics}, pages 527--536, Florence, Italy. Association for Computational Linguistics.

\bibitem[{Post(2018)}]{post2018call}
Matt Post. 2018.
\newblock \href {https://doi.org/10.18653/v1/W18-6319} {A call for clarity in reporting {BLEU} scores}.
\newblock In \emph{Proceedings of the Third Conference on Machine Translation: Research Papers}, pages 186--191, Brussels, Belgium. Association for Computational Linguistics.

\bibitem[{Radford et~al.(2023)Radford, Kim, Xu, Brockman, McLeavey, and Sutskever}]{radford2023robust}
Alec Radford, Jong~Wook Kim, Tao Xu, Greg Brockman, Christine McLeavey, and Ilya Sutskever. 2023.
\newblock \href {https://proceedings.mlr.press/v202/radford23a/radford23a.pdf} {Robust speech recognition via large-scale weak supervision}.
\newblock In \emph{International Conference on Machine Learning}, pages 28492--28518. PMLR.

\bibitem[{Rubenstein et~al.(2023)Rubenstein, Asawaroengchai, Nguyen, Bapna, Borsos, Quitry, Chen, Badawy, Han, Kharitonov et~al.}]{rubenstein2023audiopalm}
Paul~K Rubenstein, Chulayuth Asawaroengchai, Duc~Dung Nguyen, Ankur Bapna, Zal{\'a}n Borsos, F{\'e}lix de~Chaumont Quitry, Peter Chen, Dalia~El Badawy, Wei Han, Eugene Kharitonov, et~al. 2023.
\newblock \href {https://arxiv.org/abs/2306.12925} {Audiopalm: A large language model that can speak and listen}.
\newblock \emph{ArXiv preprint}, abs/2306.12925.

\bibitem[{Tang et~al.(2023)Tang, Yu, Sun, Chen, Tan, Li, Lu, Ma, and Zhang}]{tang2023salmonn}
Changli Tang, Wenyi Yu, Guangzhi Sun, Xianzhao Chen, Tian Tan, Wei Li, Lu~Lu, Zejun Ma, and Chao Zhang. 2023.
\newblock \href {https://arxiv.org/abs/2310.13289} {Salmonn: Towards generic hearing abilities for large language models}.
\newblock \emph{ArXiv preprint}, abs/2310.13289.

\bibitem[{Taori et~al.(2023)Taori, Gulrajani, Zhang, Dubois, Li, Guestrin, Liang, and Hashimoto}]{taori2023alpaca}
Rohan Taori, Ishaan Gulrajani, Tianyi Zhang, Yann Dubois, Xuechen Li, Carlos Guestrin, Percy Liang, and Tatsunori~B Hashimoto. 2023.
\newblock \href {https://crfm.stanford.edu/2023/03/13/alpaca.html} {Alpaca: A strong, replicable instruction-following model}.
\newblock \emph{Stanford Center for Research on Foundation Models}, 3(6):7.

\bibitem[{Touvron et~al.(2023{\natexlab{a}})Touvron, Lavril, Izacard, Martinet, Lachaux et~al.}]{llama}
Hugo Touvron, Thibaut Lavril, Gautier Izacard, Xavier Martinet, Marie-Anne Lachaux, et~al. 2023{\natexlab{a}}.
\newblock \href {https://arxiv.org/abs/2302.13971} {{LLaMA}: Open and efficient foundation language models}.
\newblock \emph{ArXiv preprint}, abs/2302.13971.

\bibitem[{Touvron et~al.(2023{\natexlab{b}})Touvron, Martin, Stone, Albert, Almahairi, Babaei, Bashlykov, Batra, Bhargava, Bhosale et~al.}]{touvron2023llama}
Hugo Touvron, Louis Martin, Kevin Stone, Peter Albert, Amjad Almahairi, Yasmine Babaei, Nikolay Bashlykov, Soumya Batra, Prajjwal Bhargava, Shruti Bhosale, et~al. 2023{\natexlab{b}}.
\newblock \href {https://arxiv.org/abs/2307.09288} {Llama 2: Open foundation and fine-tuned chat models}.
\newblock \emph{ArXiv preprint}, abs/2307.09288.

\bibitem[{Van~der Maaten and Hinton(2008)}]{van2008visualizing}
Laurens Van~der Maaten and Geoffrey Hinton. 2008.
\newblock \href {https://www.jmlr.org/papers/volume9/vandermaaten08a/vandermaaten08a.pdf?fbcl} {Visualizing data using t-sne.}
\newblock \emph{Journal of machine learning research}, 9(11).

\bibitem[{Vaswani et~al.(2017)Vaswani, Shazeer, Parmar, Uszkoreit, Jones, Gomez, Kaiser, and Polosukhin}]{vaswani2017attention}
Ashish Vaswani, Noam Shazeer, Niki Parmar, Jakob Uszkoreit, Llion Jones, Aidan~N. Gomez, Lukasz Kaiser, and Illia Polosukhin. 2017.
\newblock \href {https://proceedings.neurips.cc/paper/2017/hash/3f5ee243547dee91fbd053c1c4a845aa-Abstract.html} {Attention is all you need}.
\newblock In \emph{Advances in Neural Information Processing Systems 30: Annual Conference on Neural Information Processing Systems 2017, December 4-9, 2017, Long Beach, CA, {USA}}, pages 5998--6008.

\bibitem[{Wang et~al.(2020)Wang, Wu, and Pino}]{wang2020covost}
Changhan Wang, Anne Wu, and Juan Pino. 2020.
\newblock \href {https://arxiv.org/abs/2007.10310} {Covost 2 and massively multilingual speech-to-text translation}.
\newblock \emph{ArXiv preprint}, abs/2007.10310.

\bibitem[{Wang et~al.(2023{\natexlab{a}})Wang, Liao, Huang, Lu, Wu, Liu, Zong, and Zhang}]{wang2023blsp}
Chen Wang, Minpeng Liao, Zhongqiang Huang, Jinliang Lu, Junhong Wu, Yuchen Liu, Chengqing Zong, and Jiajun Zhang. 2023{\natexlab{a}}.
\newblock \href {https://arxiv.org/abs/2309.00916} {Blsp: Bootstrapping language-speech pre-training via behavior alignment of continuation writing}.
\newblock \emph{ArXiv preprint}, abs/2309.00916.

\bibitem[{Wang et~al.(2023{\natexlab{b}})Wang, Li, Dai, Chen, Zhou, Meng, Zhou, and Sun}]{wang2023label}
Lean Wang, Lei Li, Damai Dai, Deli Chen, Hao Zhou, Fandong Meng, Jie Zhou, and Xu~Sun. 2023{\natexlab{b}}.
\newblock \href {https://arxiv.org/abs/2305.14160} {Label words are anchors: An information flow perspective for understanding in-context learning}.
\newblock \emph{arXiv e-prints}, pages arXiv--2305.

\bibitem[{Wang et~al.(2023{\natexlab{c}})Wang, Zhou, Zhang, Wu, Liu, Gaur, Chen, Li, and Wei}]{wang2023viola}
Tianrui Wang, Long Zhou, Ziqiang Zhang, Yu~Wu, Shujie Liu, Yashesh Gaur, Zhuo Chen, Jinyu Li, and Furu Wei. 2023{\natexlab{c}}.
\newblock \href {https://arxiv.org/abs/2305.16107} {Viola: Unified codec language models for speech recognition, synthesis, and translation}.
\newblock \emph{ArXiv preprint}, abs/2305.16107.

\bibitem[{Wang et~al.(2023{\natexlab{d}})Wang, Ivison, Dasigi, Hessel, Khot, Chandu, Wadden, MacMillan, Smith, Beltagy et~al.}]{wang2023far}
Yizhong Wang, Hamish Ivison, Pradeep Dasigi, Jack Hessel, Tushar Khot, Khyathi Chandu, David Wadden, Kelsey MacMillan, Noah~A Smith, Iz~Beltagy, et~al. 2023{\natexlab{d}}.
\newblock \href {https://proceedings.neurips.cc/paper_files/paper/2023/file/ec6413875e4ab08d7bc4d8e225263398-Paper-Datasets_and_Benchmarks.pdf} {How far can camels go? exploring the state of instruction tuning on open resources}.
\newblock \emph{Advances in Neural Information Processing Systems}, 36:74764--74786.

\bibitem[{Wang et~al.(2023{\natexlab{e}})Wang, Kordi, Mishra, Liu, Smith, Khashabi, and Hajishirzi}]{wang2023self}
Yizhong Wang, Yeganeh Kordi, Swaroop Mishra, Alisa Liu, Noah~A Smith, Daniel Khashabi, and Hannaneh Hajishirzi. 2023{\natexlab{e}}.
\newblock \href {https://arxiv.org/abs/2212.10560} {Self-instruct: Aligning language models with self-generated instructions}.
\newblock In \emph{The 61st Annual Meeting Of The Association For Computational Linguistics}.

\bibitem[{Wiegreffe and Pinter(2019)}]{wiegreffe2019attention}
Sarah Wiegreffe and Yuval Pinter. 2019.
\newblock \href {https://doi.org/10.18653/v1/D19-1002} {Attention is not not explanation}.
\newblock In \emph{Proceedings of the 2019 Conference on Empirical Methods in Natural Language Processing and the 9th International Joint Conference on Natural Language Processing (EMNLP-IJCNLP)}, pages 11--20, Hong Kong, China. Association for Computational Linguistics.

\bibitem[{Wu et~al.(2023{\natexlab{a}})Wu, Chang, Wu, and Lee}]{wu2023speechgen}
Haibin Wu, Kai-Wei Chang, Yuan-Kuei Wu, and Hung-yi Lee. 2023{\natexlab{a}}.
\newblock \href {https://arxiv.org/abs/2306.02207} {Speechgen: Unlocking the generative power of speech language models with prompts}.
\newblock \emph{ArXiv preprint}, abs/2306.02207.

\bibitem[{Wu et~al.(2023{\natexlab{b}})Wu, Gaur, Chen, Zhou, Zhu, Wang, Li, Liu, Ren, Liu et~al.}]{wu2023decoder}
Jian Wu, Yashesh Gaur, Zhuo Chen, Long Zhou, Yimeng Zhu, Tianrui Wang, Jinyu Li, Shujie Liu, Bo~Ren, Linquan Liu, et~al. 2023{\natexlab{b}}.
\newblock \href {https://arxiv.org/abs/2307.03917} {On decoder-only architecture for speech-to-text and large language model integration}.
\newblock \emph{ArXiv preprint}, abs/2307.03917.

\bibitem[{Xu et~al.(2024)Xu, Sun, Zheng, Geng, Zhao, Feng, Tao, Lin, and Jiang}]{xu2024wizardlm}
Can Xu, Qingfeng Sun, Kai Zheng, Xiubo Geng, Pu~Zhao, Jiazhan Feng, Chongyang Tao, Qingwei Lin, and Daxin Jiang. 2024.
\newblock \href {https://openreview.net/forum?id=CfXh93NDgH} {Wizardlm: Empowering large pre-trained language models to follow complex instructions}.
\newblock In \emph{The Twelfth International Conference on Learning Representations}.

\bibitem[{Xu et~al.(2023)Xu, Kim, Sharaf, and Awadalla}]{xu2023paradigm}
Haoran Xu, Young~Jin Kim, Amr Sharaf, and Hany~Hassan Awadalla. 2023.
\newblock \href {https://arxiv.org/abs/2309.11674} {A paradigm shift in machine translation: Boosting translation performance of large language models}.
\newblock \emph{ArXiv preprint}, abs/2309.11674.

\bibitem[{Yang et~al.(2024)Yang, Pang, Feng, Wang, Chen, Zhu, and Liu}]{yang2024self}
Zhaorui Yang, Tianyu Pang, Haozhe Feng, Han Wang, Wei Chen, Minfeng Zhu, and Qian Liu. 2024.
\newblock \href {https://arxiv.org/abs/2402.13669} {Self-distillation bridges distribution gap in language model fine-tuning}.
\newblock \emph{arXiv preprint arXiv:2402.13669}.

\bibitem[{Yu et~al.(2023{\natexlab{a}})Yu, Ding, Liu, Chen, Zhang, Tao, and Zhang}]{yu-etal-2023-promptst}
Tengfei Yu, Liang Ding, Xuebo Liu, Kehai Chen, Meishan Zhang, Dacheng Tao, and Min Zhang. 2023{\natexlab{a}}.
\newblock \href {https://doi.org/10.18653/v1/2023.emnlp-main.627} {{P}rompt{ST}: Abstract prompt learning for end-to-end speech translation}.
\newblock In \emph{Proceedings of the 2023 Conference on Empirical Methods in Natural Language Processing}, pages 10140--10154, Singapore. Association for Computational Linguistics.

\bibitem[{Yu et~al.(2023{\natexlab{b}})Yu, Tang, Sun, Chen, Tan, Li, Lu, Ma, and Zhang}]{yuwenyi}
Wenyi Yu, Changli Tang, Guangzhi Sun, Xianzhao Chen, Tian Tan, Wei Li, Lu~Lu, Zejun Ma, and Chao Zhang. 2023{\natexlab{b}}.
\newblock \href {https://arxiv.org/abs/2309.13963} {Connecting speech encoder and large language model for {ASR}}.
\newblock \emph{ArXiv preprint}, abs/2309.13963.

\bibitem[{Yuan et~al.(2024)Yuan, Pang, Cho, Sukhbaatar, Xu, and Weston}]{yuan2024self}
Weizhe Yuan, Richard~Yuanzhe Pang, Kyunghyun Cho, Sainbayar Sukhbaatar, Jing Xu, and Jason Weston. 2024.
\newblock \href {https://arxiv.org/abs/2401.10020} {Self-rewarding language models}.
\newblock \emph{arXiv preprint arXiv:2401.10020}.

\bibitem[{Zhang et~al.(2023{\natexlab{a}})Zhang, Li, Zhang, Zhan, Wang, Zhou, and Qiu}]{zhang2023speechgpt}
Dong Zhang, Shimin Li, Xin Zhang, Jun Zhan, Pengyu Wang, Yaqian Zhou, and Xipeng Qiu. 2023{\natexlab{a}}.
\newblock \href {https://arxiv.org/abs/2305.11000} {Speechgpt: Empowering large language models with intrinsic cross-modal conversational abilities}.
\newblock \emph{ArXiv preprint}, abs/2305.11000.

\bibitem[{Zhang et~al.(2023{\natexlab{b}})Zhang, Li, and Bing}]{zhang2023video}
Hang Zhang, Xin Li, and Lidong Bing. 2023{\natexlab{b}}.
\newblock \href {https://arxiv.org/abs/2306.02858} {Video-llama: An instruction-tuned audio-visual language model for video understanding}.
\newblock \emph{arXiv e-prints}, pages arXiv--2306.

\bibitem[{Zhang et~al.(2019)Zhang, Kishore, Wu, Weinberger, and Artzi}]{zhang2019bertscore}
Tianyi Zhang, Varsha Kishore, Felix Wu, Kilian~Q Weinberger, and Yoav Artzi. 2019.
\newblock \href {https://arxiv.org/abs/1904.09675} {Bertscore: Evaluating text generation with bert}.
\newblock In \emph{International Conference on Learning Representations}.

\bibitem[{Zhang et~al.(2023{\natexlab{c}})Zhang, Zhang, Li, Zhou, and Qiu}]{zhang2023speechtokenizer}
Xin Zhang, Dong Zhang, Shimin Li, Yaqian Zhou, and Xipeng Qiu. 2023{\natexlab{c}}.
\newblock \href {https://arxiv.org/abs/2308.16692} {Speechtokenizer: Unified speech tokenizer for speech large language models}.
\newblock \emph{ArXiv preprint}, abs/2308.16692.

\bibitem[{Zhou et~al.(2023)Zhou, Liu, Xu, Iyer, Sun, Mao, Ma, Efrat, Yu, Yu et~al.}]{zhou2023lima}
Chunting Zhou, Pengfei Liu, Puxin Xu, Srini Iyer, Jiao Sun, Yuning Mao, Xuezhe Ma, Avia Efrat, Ping Yu, Lili Yu, et~al. 2023.
\newblock \href {https://arxiv.org/abs/2305.11206} {Lima: Less is more for alignment}.
\newblock \emph{ArXiv preprint}, abs/2305.11206.

\end{thebibliography}
\clearpage
\appendix

\section{Appendix}
\subsection{Attention as Explanation}
\begin{table}[t]
\centering
\scalebox{0.75}{
\begin{tabular}{l cc cc }
\toprule
\multirow{2}{*}{\bf Model} &  \multicolumn{2}{c}{\bf CoVoST} &\multicolumn{2}{c}{\bf MuST-C} \\
\cmidrule(lr){2-3} \cmidrule(lr){4-5}
 &  en-zh & en-de & en-es & en-fr \\ 
\midrule
Qwen-Audio & \bf 30.8 &\bf 18.2& 20.7&23.2 \\ 
Vanilla IT   & 0.1& 0.2& 0.0 & 0.0 \\ 
BLSP*    & 1.3 & 13.7 & 12.2 & 17.0 \\ 
BLSP   & 10.8 & 6.7 & 14.8 & 14.5 \\ 
Self-Powered LSM Small   & 20.2 & 14.8& 18.9 & 23.9 \\ 
Self-Powered LSM Medium   & 22.7 &  16.2 & \bf 21.1 & 26.3 \\ 
Self-Powered LSM Large   & 23.9 & 17.4& 20.7 & \bf 26.8 \\ 

\bottomrule
\end{tabular}}
\caption{CoVoST-2 and MuST-C ST results.}
\label{tab:st}
\end{table}

\begin{table*}[htbp]
\begin{tcolorbox}
{\bf Speech Translation}\\ 
~~Provide the translation text from English to Chinese according to the speech.  \\
{\bf Speech Recognition}  \\
~~Provide the transcription according to the speech. \\
{\bf Speech Summarization}  \\
~~Listen to the speech and provide a summary, capturing the main points in no more than three sentences.    \\
{\bf Keyword Extraction}  \\
~~Identify and list the keywords or phrases from the speech, focusing on the most relevant terms used.   \\
{\bf Question Generation} \\
~~Based on the content of the speech, generate a relevant question. 
\end{tcolorbox}
\caption{Instructions used for attention and representation analysis}
\label{tab:attention_vis}
\end{table*}

\begin{table*}[htbp]
\begin{tcolorbox}
{\bf Speech Recognition}\\ 
Provide the transcription according to the speech.\\
{\bf Speech Translation}  \\
Provide the translation text from \{source\} to \{target\} according to the speech. (Do not generate extra information) \\
{\bf Emotion Recognition}  \\
 Classify the emotion of the speech from \{'neutral', 'joy', 'sadness', 'anger', 'surprise', 'fear', 'disgust'\}. Ensure your response strictly adheres to this format: \{'xxx'\}. \\
{\bf Intent Recognition}  \\
Classify one of the intent label in ['activate lamp', 'activate lights', 'activate music', 'bring juice', 'bring newspaper', 'bring shoes', 'bring socks', 'change language Chinese', 'change language English', 'change language German', 'change language Korean', 'change language none', 'deactivate lamp', 'deactivate lights', 'deactivate music', 'decrease heat', 'decrease volume', 'increase heat', 'increase volume']  according to the speech.\\
{\bf Keyword Extraction - Light}  \\
Please listen carefully to the SPEECH provided and extract two keywords from the following list: {'bedroom', 'brightness', 'decrease', 'increase', 'kitchen', 'living room', 'turn off', 'turn on'}. Your response should strictly follow this format: ['keyword1', 'keyword2']. 
\\
{\bf Keyword Extraction - Water}  \\
Please listen carefully to the SPEECH provided and extract three keywords from the following list: {'bedroom', 'brightness', 'decrease', 'increase', 'kitchen', 'living room', 'turn off', 'turn on'}. Your response should strictly follow this format: ['keyword1', 'keyword2', 'keyword3'].   \\
{\bf BoolQA}  \\
Answer the questions in speech based on the CONTEXT given,your answer is only true or false, you don’t need to answer anything else. CONTEXT: \{references\} \\
{\bf WebQA}  \\
Answer the questions in speech based on the CONTEXT given: CONTEXT: \{references\}
\end{tcolorbox}
\caption{Instructions used for evaluation.}
\label{appendix:evaluate}
\end{table*} 
\paragraph{Formulation} 
\label{Formulation}
Attention weights are a widely used interpretation tool that identifies patterns in attention interactions to determine which input vectors are most influential.
Each self-attention weight $\alpha_{i,j}$ is computed from the corresponding input $\mathbf{X}=[\mathbf{x}_i]^I_{i=1}\in\mathbb{R}^{I\times d}$:

{
\footnotesize
\begin{equation}
    \alpha_{i,j} = \mathop{softmax}_{\mathbf{x}_j \in \mathbf{X}}\left(\frac{\mathop{q(\mathbf{x}_i)} \mathop{k(\mathbf{x}_j)^\top}}{\sqrt{d}}\right)\in\mathbb{R} ,
    \label{eq2:attention_weight1}
\end{equation}}

where $\mathop{q(\cdot)}$, $\mathop{k(\cdot)}$ are the query and key transformations, respectively. For multi-modal inputs, the features themselves also affect critical token interactions. 
We define the \textit{norm of the weighted transformed vector} as $\mathbf{a}(i,j)$, which compute from the self-attention weight $\alpha_{i,j}$ and the $j$-th input vector $\mathbf{x}_j$:

{\footnotesize
\begin{equation}
    \label{eq:final-loss2}
     \mathbf{a}(i,j) =\lVert\alpha_{i,j} \left(\mathop{v(\mathbf{x}_j)} \mathbf{W}^O\right)\rVert ,
\end{equation}}

where $\mathop{v(\cdot)}$ is value transformations, $\mathbf{W}^O$ was introduced in~\citet{vaswani2017attention} which integrate the output of multi-heads applied to each attention. 
Here, $\mathbf{a}(i,j)$ can represent the norm-based attention score from the $j$-th word to the $i$-th word. When the model generates a $M$-sized text for every layer, we compute the average attention scores $\mathbf{A}$:

{
\footnotesize
\begin{equation}
    \label{eq:final-loss3}
    \begin{split}
        \mathbf{A}_j = \frac{1}{M} \sum_{m=1}^{M}\mathbf{a}(m,j) .
    \end{split}
\end{equation}}

Here, $\mathbf{a}(m,j)$ represents the attention score from the $j$-th token to the $m$-th generated token, with $\mathbf{A}_j$ denoting the average attention score from the $j$-th token to the model's output.
\paragraph{Experiments Setup} 
\label{setup}
We conduct a comparative analysis of layer-wise behavior between LLMs and LSMs regarding their handling of instruction and source inputs. For LLMs, our analysis includes Llama2-base \citep{touvron2023llama}, Vicuna-1.5-7B \citep{chiang2023vicuna}, and Alpaca-7B \citep{taori2023alpaca}. For LSMs, we evaluate an untrained model, a model trained on the ASR task using the LibriSpeech-960h dataset, and a model trained on En-Zh ST task using the CoVoST2 dataset. Each model comprises $32$ layers, which we categorize and average into six levels for enhanced visualization clarity.

\paragraph{Instructions Used for Visualization}
\label{appendix:visual}
The instructions used for attention and speech-text representation visualization are presented in Table~\ref{tab:attention_vis}. To enhance the reliability of our analysis, we diversify it by including five distinct instruction types, covering a broad spectrum of tasks: speech translation, automatic speech recognition, speech summarization, keyword extraction, and question generation. For each of these categories, we randomly select 240 samples from three different datasets:  LibriSpeech test set, the CoVoST2 En-Zh test set, and the MuST-C En-Zh tst-COMMON dataset, resulting in a total test set of 1,200 samples.

\subsection{Details of Experimental Results}
\paragraph{Detail ST Results}
\label{appendix:main}
Table~\ref{tab:st} showcases the outcomes of our ST results for a range of language pairs, including CoVoST En-\{De, Zh\}, as well as MUST-C English-to-\{Es, Fr\} ST tasks. The result shows that the Self-Powered LSM achieves comparable performance to existing baseline, confirming the efficacy of our method.

\subsection{Instructions for Self-Powered LSM}
\label{appendix:three}
\begin{table*}[t]
\begin{tcolorbox}

{\bf Content Repetition}\\ 
1. Repeat the provided text, ensuring to maintain its original meaning and details

2. Rephrase the text without altering its initial intent and key information

3. Paraphrase the provided text while preserving all original facts and nuances

4. Echo the content of the text, maintaining its exact purpose and details

5. Retell the given information without changing its meaning or losing any critical data

{\bf Translation}  \\
1. Provide the translation from English to \{target\}

2. Translate the given English content into \{target\}

3. Render the English text into \{target\}

4. Convert the specified English text into \{target\}

5. Translate the provided English material into the \{target\} language\\
{\bf Keyword Extraction}  \\
1. Extract the most frequently occurring words or phrases in the text, excluding common stopwords, to identify main topics

2. Identify and list the most common words or phrases from the text, omitting typical stopwords, to highlight central themes

3. Determine the key words or phrases frequently used in the text, removing all usual stopwords, to discern the main topics

4. Find the recurring words or phrases in the text, ignoring common stopwords, to ascertain the primary themes

5. Extract significant words or phrases that appear often in the text, exclude basic stopwords, to uncover the main subjects

{\bf Intent Recognition}  \\
1. Determine the primary purpose of the speech and evaluate how clearly and effectively the message is conveyed

2. Identify the main intent of the speech and assess the clarity and effectiveness of its delivery

3. Ascertain the fundamental objective of the speech and critique the transparency and efficiency of its presentation

4. Figure out the aim of the speech and judge how lucidly and effectively the ideas are presented

5. Assess the central purpose of the speech and evaluate the directness and impact of its expression

{\bf Sentiment Analysis}  \\
1. Determine the sentiment of the text and identify which sections contribute most to sentiment

2. Analyze the overall mood of the text and pinpoint the parts that heavily influence the sentiment

3. Evaluate the emotional tone of the text and determine which segments primarily affect the sentiment

4. Identify the feeling conveyed by the text and specify which portions substantially shape this sentiment

5. Assess the sentiment expressed in the text and highlight which areas contribute most to this feeling

{\bf Continuation}  \\
1. Please write a coherent and engaging English continuation of the given English text with less than 50 words

2. Compose a logical and captivating follow-up to the provided English text within 50 words

3. Craft a coherent extension for the English text, ensuring it does not exceed 50 words

4. Develop a consistent and attractive continuation of the English text, keeping it under 50 words

5. Write a fluent and engaging continuation of the English text, limited to 50 words\\
\end{tcolorbox}
\caption{Instructions used for data generation.}
\label{tab:genration_template}
\end{table*}

\begin{table*}[t]
\begin{tcolorbox}
{\bf Speech Recognition}\\ 
1. Provide the English transcription according to the speech \\
2. Transcribe the spoken words into written English text \\
3. Convert the spoken English into a written transcript \\
4. Create a written transcription of the spoken English \\
5. Write down the English speech as a text transcript \\
\end{tcolorbox}
\caption{Instructions used for ground-truth ASR datasets.}
\label{tab:genration_template1}
\end{table*} 

\paragraph{Instructions Used for Data Generation}
\label{appendix:generation}

The instructions utilized for data generation are detailed in Table~\ref{tab:genration_template}. We employ six distinct types of tasks spanning various tasks: content repetition, keyword extraction, intent recognition, sentiment analysis, continuation, and translation. These instructions are used to prompt LLM to generate self-powered data and to expand the speech ability of LLM. For the ground-truth ASR dataset, the instructions we used are presented in Table~\ref{tab:genration_template1}.

\paragraph{Instructions Used for Evaluation}
\label{appendix:evaluation}
The instructions used for evaluation are presented in Table~\ref{appendix:evaluate}. For all LSMs evaluated, we use the same instructions to ensure a fair comparison.

\end{document}